\def\confName{CVPR}
\def\confYear{2026}

\def\paperTitle{3DReflecNet: A Large-Scale Dataset for 3D Reconstruction of Reflective, Transparent, and Low-Texture Objects}

\def\authorBlock{
    Zhicheng Liang$^{1}$ \quad
    Haoyi Yu$^{1}$ \quad
    Boyan Li$^{1}$ \quad
    Dayou Zhang$^{2}$ \quad
    Zijian Cao$^{1}$ \quad
    Tianyi Gong$^{1}$ \quad
    Junhua Liu$^{3}$ \\
    Shuguang Cui$^{1}$ \quad
    Fangxin Wang$^{1}$\thanks{Corresponding Author.} \\
    
    $^{1}$The Chinese University of Hong Kong, Shenzhen \\
    $^{2}$Capital Normal University \quad
    $^{3}$University of Southern California \\
    
    {\tt\small \{zhichengliang1, haoyiyu, boyanli, zijiancao, tianyigong\}@link.cuhk.edu.cn} \\
    \vspace{0.5em}
    {\tt\small zhangdayou@cnu.edu.cn, junhua.liu.0@usc.edu, \{shuguangcui, wangfangxin\}@cuhk.edu.cn}
}

\newif\ifreview 
\newif\ifarxiv \newcommand{\arxiv}{\arxivtrue}
\newif\ifcamera 
\newif\ifrebuttal 

\arxiv 

\pdfoutput=1

\documentclass[10pt,twocolumn,letterpaper]{article}

\ifreview \usepackage[review]{cvpr} \fi
\ifarxiv \usepackage[pagenumbers]{cvpr} \fi
\ifrebuttal \usepackage[rebuttal]{cvpr} \fi
\ifcamera \usepackage{cvpr} \fi

\usepackage{graphicx}	
\usepackage{amsmath}	
\usepackage{amssymb}	
\usepackage{booktabs}
\usepackage{times}
\usepackage{microtype}
\usepackage{epsfig}
\usepackage{caption}
\usepackage{float}
\usepackage{placeins}
\usepackage{color, colortbl}
\usepackage{stfloats}
\usepackage{enumitem}
\usepackage{tabularx}
\usepackage{xstring}
\usepackage{multirow}
\usepackage{xspace}
\usepackage{url}
\usepackage{subcaption}
\usepackage{xcolor}
\usepackage[hang,flushmargin]{footmisc}

\ifcamera \usepackage[accsupp]{axessibility} \fi

% ...

\ifarxiv  \fi

\newcommand{\R}[1]{{%
    \textbf{%
        \ifstrequal{#1}{1}{\textcolor{red}{R#1}}{%
        \ifstrequal{#1}{2}{\textcolor{blue}{R#1}}{%
        \ifstrequal{#1}{3}{\textcolor{magenta}{R#1}}{%
        \ifstrequal{#1}{4}{\textcolor{teal}{R#1}}{%
                           \textcolor{cyan}{R#1}%
        }}}}%
    }%
}}

\usepackage{xr-hyper}

\makeatletter
\newcommand*{\addFileDependency}[1]{
  \typeout{(#1)}
  \@addtofilelist{#1}
  \IfFileExists{#1}{}{\typeout{No file #1.}}
}

\makeatother

\definecolor{cvprblue}{rgb}{0.21,0.49,0.74}
\usepackage[pagebackref,breaklinks,colorlinks,allcolors=cvprblue]{hyperref}
\usepackage[capitalize]{cleveref}
\crefname{section}{Sec.}{Secs.}
\crefname{table}{Table}{Tables}
\crefname{figure}{Fig.}{Figs.}

\ifarxiv \crefname{appendix}{App.}{Apps.}
\else \crefname{appendix}{Suppl.}{Suppls.} \fi

\usepackage{algorithm}
\usepackage{algpseudocode}
\usepackage{amssymb}
\usepackage{amsmath}
\usepackage{xcolor}
\usepackage{comment}
\usepackage{graphicx}

\usepackage{booktabs}
\usepackage{multirow}

\usepackage{tikz} 
\usepackage{subcaption}

\usepackage{lipsum}
\usepackage{graphicx}
\usepackage{wrapfig}
\usepackage{adjustbox} 
\usepackage{colortbl}
\usepackage{cuted}
\usepackage{capt-of}
\usepackage{xspace}

\usepackage{appendix} 
\usepackage{titletoc}

\usepackage{pifont}

\newcommand{\cmark}{\ding{51}}
\newcommand{\xmark}{\ding{55}}

\usepackage{etoolbox}

\newcommand{\method}{\text{3DReflecNet}\xspace}

\newcommand{\numsyn}{12,000+\xspace}

\newcommand{\nummaterials}{22\xspace}
\newcommand{\numscenes}{2700+\xspace}

\usepackage[most]{tcolorbox}
\usepackage{float}
\usepackage{xspace}
\tcbset{
  aibox/.style={
    width=474.18663pt,
    top=10pt,
    colback=white,
    colframe=black,
    colbacktitle=black,
    enhanced,
    center,
    attach boxed title to top left={yshift=-0.1in,xshift=0.15in},
    boxed title style={boxrule=0pt,colframe=white,},
  }
}
\newtcolorbox{AIbox}[2][]{aibox,title=#2,#1}

\unless\ifarxiv
\fi
\title{\paperTitle}
\author{\authorBlock}

\begin{document}

\makeatletter
\let\old@maketitle\@maketitle
\renewcommand{\@maketitle}{
    \setcounter{figure}{0} 
    \old@maketitle
    \vspace{-20pt}
    \begin{center}
        \includegraphics[width=\linewidth]{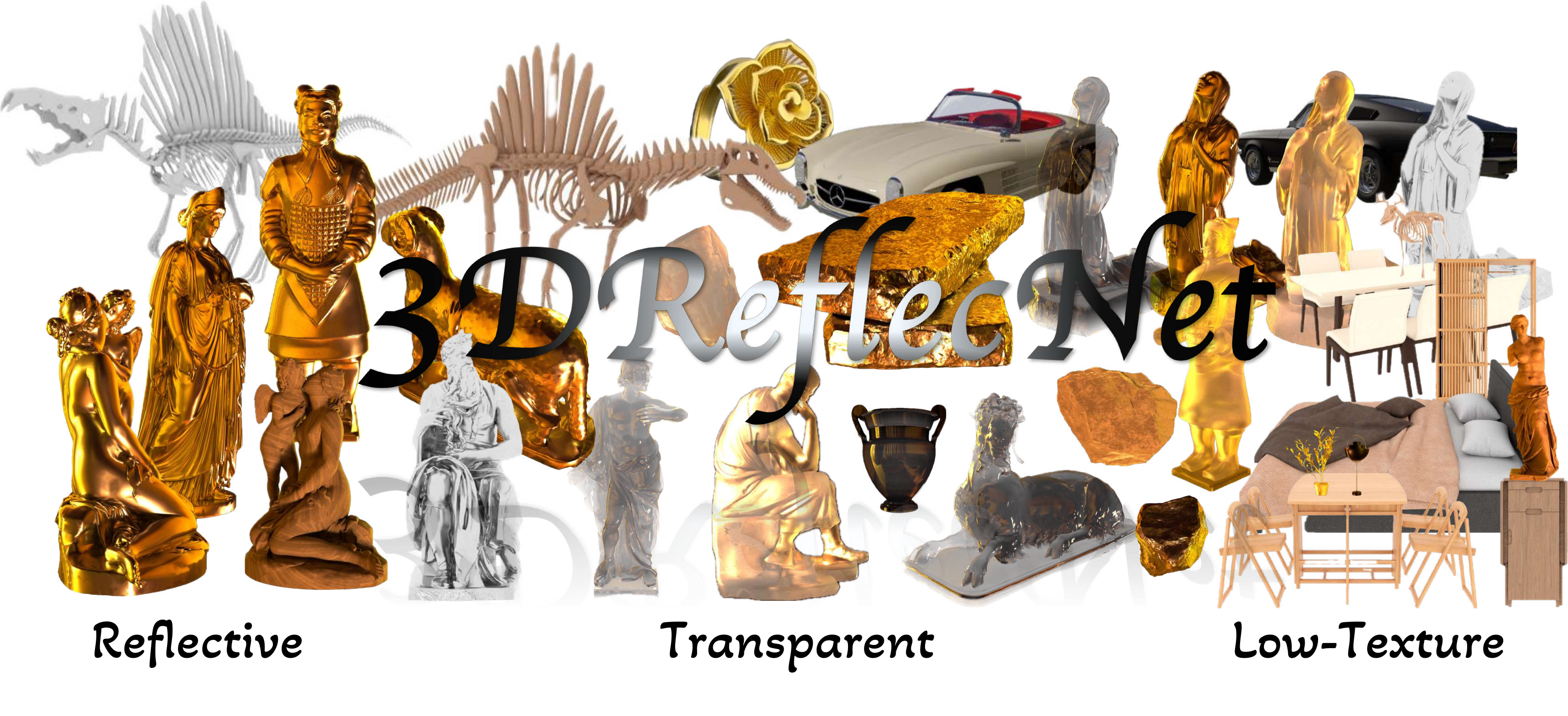}
        \vspace{-8mm}
        \captionof{figure}{\method: A large-scale multi-view, object-centric dataset featuring reflective, transparent, and low-texture objects. Providing high-quality annotations for 3D reconstruction.}
        \label{fig:teaser}
    \end{center}
}
\makeatother

\maketitle



\begin{abstract}

Accurate 3D reconstruction of objects with reflective, transparent, or low-texture surfaces still remains notoriously challenging. Such materials often violate key assumptions in multi-view reconstruction pipelines, such as photometric consistency and the availability on distinct geometric texture cues. Existing datasets primarily focus on diffuse, textured objects, and therefore provide limited insight into performance under real-world material complexities. 

We introduce \textbf{\method}, a large-scale hybrid dataset exceeding 22 TB that is specifically designed to benchmark and advance 3D vision methods for these challenging materials. \method combines two types of data: over 120,000 synthetic instances generated via physically-based rendering of more than 12,000 shapes, and over 1,000 real-world objects captured using consumer devices. Together, these data consist of more than 7 million multi-view frames. The dataset spans diverse materials, complex lighting conditions, and a wide range of geometric forms—including shapes generated from both real and LLM-synthesized 2D images using diffusion-based pipelines. To support robust evaluation, we design benchmarks for five core tasks: image matching, structure-from-motion, novel view synthesis, reflection removal, and relighting. Extensive experiments demonstrate that state-of-the-art methods struggle to maintain accuracy across these settings, highlighting the need for more resilient 3D vision models. 

\end{abstract}
\section{Introduction}

Multi-view 3D reconstruction is a central problem in computer vision, fundamental to applications spanning robotics, AR/VR, autonomous driving, and digital content creation. Recent advances such as Neural Radiance Fields (NeRF)~\cite{mildenhall2021nerf} and its variants~\cite{sun2022direct, muller2022instant, fridovich2022plenoxels}, along with more recent advancements in 3D Gaussian Splatting and related approaches~\cite{kerbl20233d, huang20242d, turkulainen2025ags, zhou2024diffgs, Yu_2024_CVPR, matsuki2024gaussian, fu2024colmap} have significantly improved reconstruction quality and rendering efficiency, particularly for textured, Lambertian surfaces. However, these methods struggle when confronted with complex optical phenomena—notably specular reflection, transparency, and low texture. Such conditions induce appearance inconsistency across views, undermining the assumptions behind Structure-from-Motion (SfM)~\cite{crandall2012sfm, cui2017hsfm, hartley2003multiple, jiang2013global} and Multi-View Stereo (MVS) pipelines~\cite{galliani2015massively, wang2023adaptive,oechsle2021unisurf, ma2022multiview}. 

\begin{figure}
    \centering
    \includegraphics[width=\linewidth]{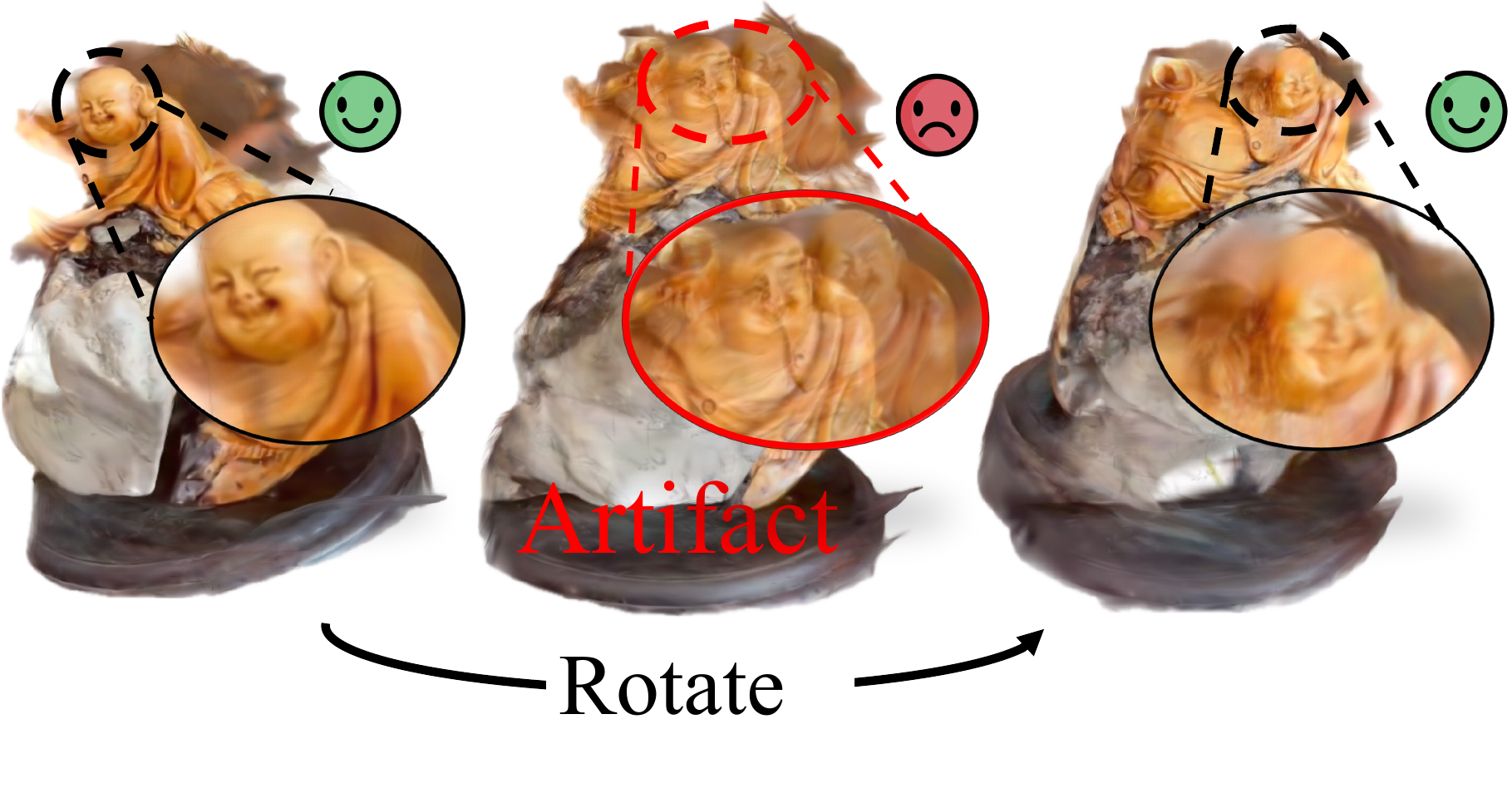}
    \vspace{-8mm}
    \caption{Inaccurate camera pose estimation leads to reconstruction artifacts.}
    \vspace{-5mm}
    \label{fig:artifact}

\end{figure}

The root cause lies in the foundational assumptions of existing multi-view reconstruction methods. Most state-of-the-art algorithms rely on (i) photometric consistency and (ii) distinctive appearance features across views. Both assumptions break down under view-dependent effects, light transmission, or low-texture surfaces. In such cases, both SfM and MVS pipelines fail to robustly recover accurate camera poses or dense geometry. Figure~\ref{fig:artifact} illustrates a typical failure mode, where even slight pose errors on reflective or transparent surfaces introduce geometry inconsistencies and rendering artifacts across viewpoints.

Despite the growing interest in robust 3D reconstruction, current benchmarks often overlook these challenging materials. Datasets such as DTU~\cite{jensen2014large}, CO3D~\cite{reizenstein2021common}, and MVImgNet~\cite{wu2023mvimgnet} largely focus on textured, diffuse objects under relatively stable lighting and capture conditions. OpenMaterial~\cite{christy2023openmaterial} makes important progress by introducing synthetic multiview images rendered with physically measured refraction indices. However, it remains confined to purely synthetic data, lacks real-world noise and motion, and primarily supports novel view synthesis and geometry evaluation under controlled assumptions.

In this paper, we present \textbf{\method}, a large-scale hybrid dataset specifically designed for the reconstruction of reflective, transparent, and low-texture objects. \method combines \numsyn physically-based rendered synthetic instances with more than 1,000 real-world scans captured using commodity devices. It features extensive material diversity (e.g., polished metal, glass, ceramics), view-dependent specular simulation (via camera-glass setups), and rare geometric forms, including diffusion-generated 3D assets from both real-world and LLM-synthesized 2D references. 

Beyond its material complexity, \method supports a wide range of challenging tasks unaddressed by existing datasets: (i) photometrically inconsistent image matching, (ii) structure-from-motion for non-Lambertian and low-texture surfaces, (iii) novel view synthesis under complex material conditions, (iv) reflection and highlight removal, and (v) object relighting. By bridging synthetic and real domains and incorporating rich photometric and geometric variation, \method provides a new foundation to evaluate 3D vision models under realistic and challenging conditions. In summary, our contributions are threefold:

\begin{itemize}
    \item We present \textbf{3DReflecNet}, a large-scale hybrid dataset focused on reflective, transparent, and low-texture objects. It contains more than 120,000 synthetic instances and 1,000 real-world captures, totaling more than 7 million frames. The dataset spans diverse geometry, materials, and lighting across nine high-level categories.

    \item We introduce a unified asset-creation pipeline that combines physically based rendering (with a camera-through-glass setup for view-dependent optics) and diffusion-driven 2D-to-3D generation, producing rich geometry and appearance diversity beyond existing libraries.

    \item We establish benchmarks for five tasks, including image matching, SfM, novel view synthesis, reflection removal, and relighting, together with standardized evaluations and baselines. Our large-scale analysis reveals systematic failure modes in state-of-the-art methods, highlighting the need for more physically aware 3D vision models.
\end{itemize}

\vspace{-7pt}

\section{Related Works}

In this section, we review prior work on 3D reconstruction and related datasets to highlight current gaps. More work on related tasks, such as specular highlight removal, reflection removal, and relighting, is discussed in Suppl.~\ref{sup-appendix:related works}.

\subsection{Multi-view 3D Reconstruction}

Image matching is fundamental to multi-view 3D reconstruction pipelines. Classical descriptors~\cite{sift, surf, orb} remain attractive for their efficiency, whereas learning-based models~\cite{superpoint,d2net, disk} improve robustness under geometric variation. More recently, detector-free approaches~\cite{loftr, edstedt2024roma, chen2022aspanformer} leverage Transformer architectures to estimate dense correspondences. However, most methods are evaluated on textured, diffuse surfaces, overlooking the cases we target.

Structure-from-Motion (SfM) estimates camera poses and sparse scene geometry via feature matching and bundle adjustment~\cite{bundle}. Modern SfM methods span incremental~\cite{incremental} and global strategies~\cite{chatterjee2013efficient, wilson2013network}, with robust extensions handling noise and outliers~\cite{wilson2014robust, ozyesil2015robust, hand2018shapefit}. Multi-View Stereo (MVS) builds dense geometry from calibrated images using both classical~\cite{schonberger2016pixelwise, galliani2015massively} and learning-based methods~\cite{wang2023adaptive, oechsle2021unisurf, ye2023constraining}. 
They rely on accurate pose estimates, making them sensitive to errors in complex materials.

Significant progress in novel view synthesis (NVS) has followed the introduction of NeRF~\cite{mildenhall2021nerf} and its many variants~\cite{sun2022direct, muller2022instant, fridovich2022plenoxels}. Recent advances in 3D Gaussian Splatting and related approaches~\cite{kerbl20233d, huang20242d, turkulainen2025ags, zhou2024diffgs, Yu_2024_CVPR, matsuki2024gaussian, fu2024colmap} have achieve state-of-the-art results. However, these methods implicitly rely on color and appearance consistency, making them less effective for reflective or transparent surfaces.

\subsection{3D Datasets and Material Complexity}

Several datasets have explored material-aware 3D reconstruction, but each addresses only a subset of the underlying challenges. MV Reflectance~\cite{oxholm2014multiview} and NeRO~\cite{liu2023nero} focus on reflective surfaces but offer limited object diversity. OpenMaterial~\cite{dang2024openmaterial} introduces 1001 synthetic objects rendered under measured optical properties, but lacks real-world data and only supports a narrow range of tasks. DTU~\cite{aanaes2016large}, Tanks and Temples~\cite{Knapitsch2017}, and BlendedMVS~\cite{yao2020blendedmvs} focus on multi-view reconstruction with known geometry but primarily feature diffuse materials. Large-scale repositories such as ShapeNet-Intrinsics~\cite{shi2017learning}, ABO~\cite{collins2022abo}, RTMV~\cite{tremblay2022rtmv}, and Objaverse~\cite{deitke2023objaverse} expand shape and appearance variation, though they often lack physically plausible material simulation. In contrast, \method combines high-fidelity synthetic rendering with real-world captures, providing both optical realism and broad task coverage across five evaluation tracks.

\section{Observations and Motivation}

\begin{figure}
    \centering
    \includegraphics[width=\linewidth]{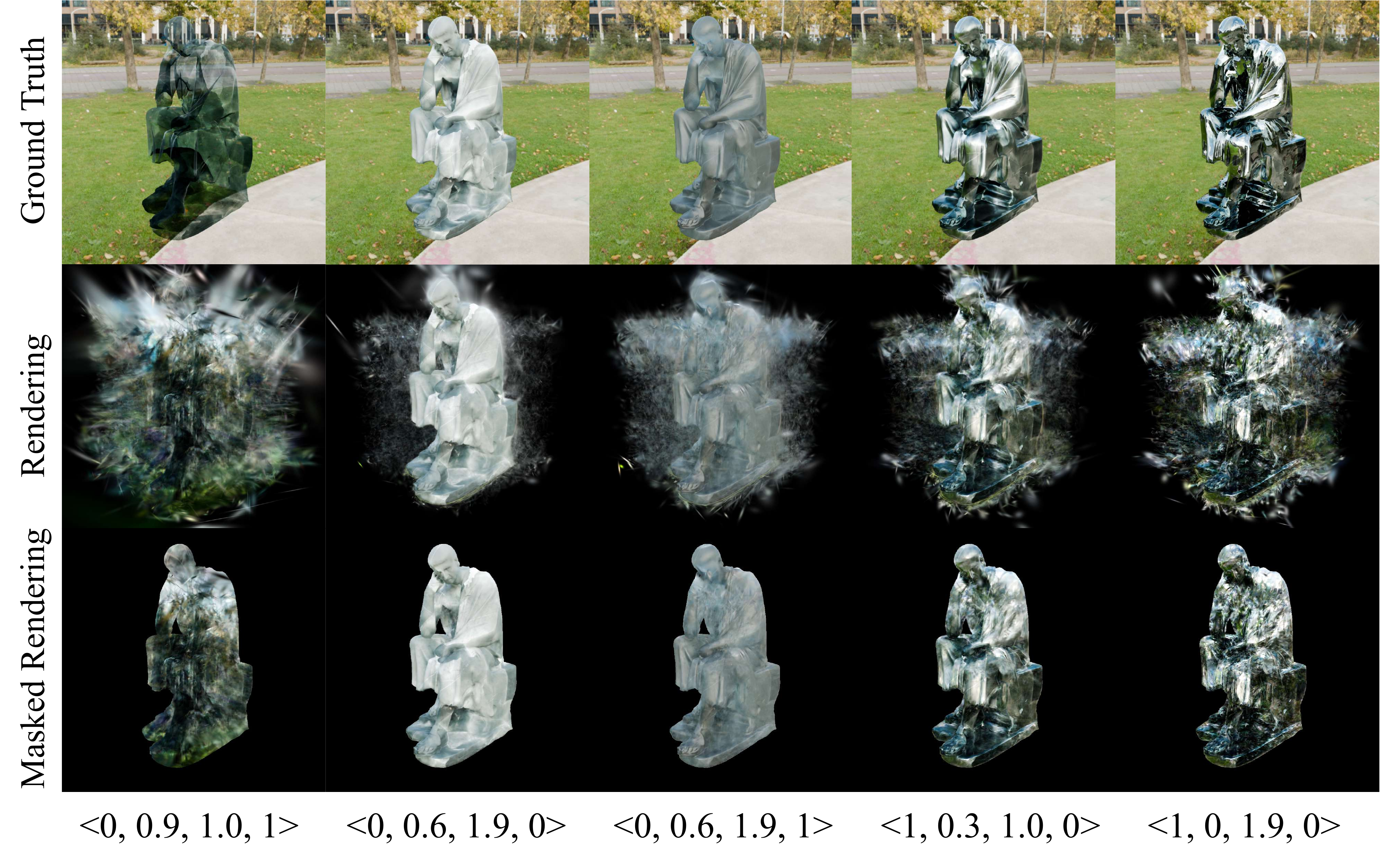}
    \vspace{-7mm}
    \caption{Effect of material properties on physically-based rendering and reconstruction. Each column shows a different material configuration $\langle$Metallic, Roughness, IOR, Transmission$\rangle$. Rows (top to bottom): sample input view, final reconstructed view, and masked reconstructed view for comparison with ground truth.}
    \vspace{-7mm}
    \label{fig:materials_training}
\end{figure}

\begin{figure}
    \centering
    \includegraphics[width=\linewidth]{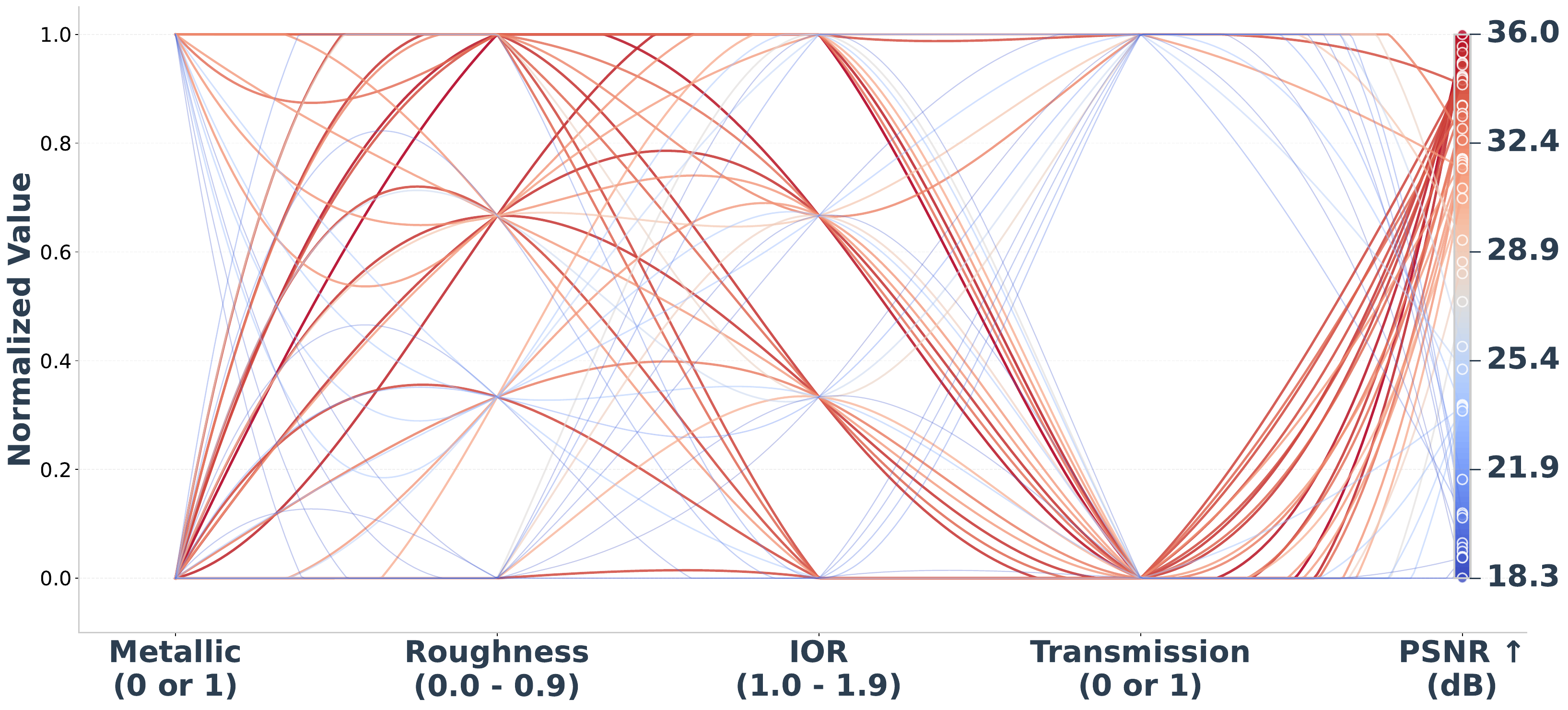}
    \vspace{-7mm}
    \caption{Material parameter sweep across 48 configurations. Each line represents a single trial, colored by reconstruction quality (PSNR). The plot demonstrates how material properties systematically affect reconstruction performance.}
    \vspace{-5mm}
    \label{fig:material_parameters}
\end{figure}

\begin{figure*}[ht!]
    \centering
    \includegraphics[width=\linewidth]{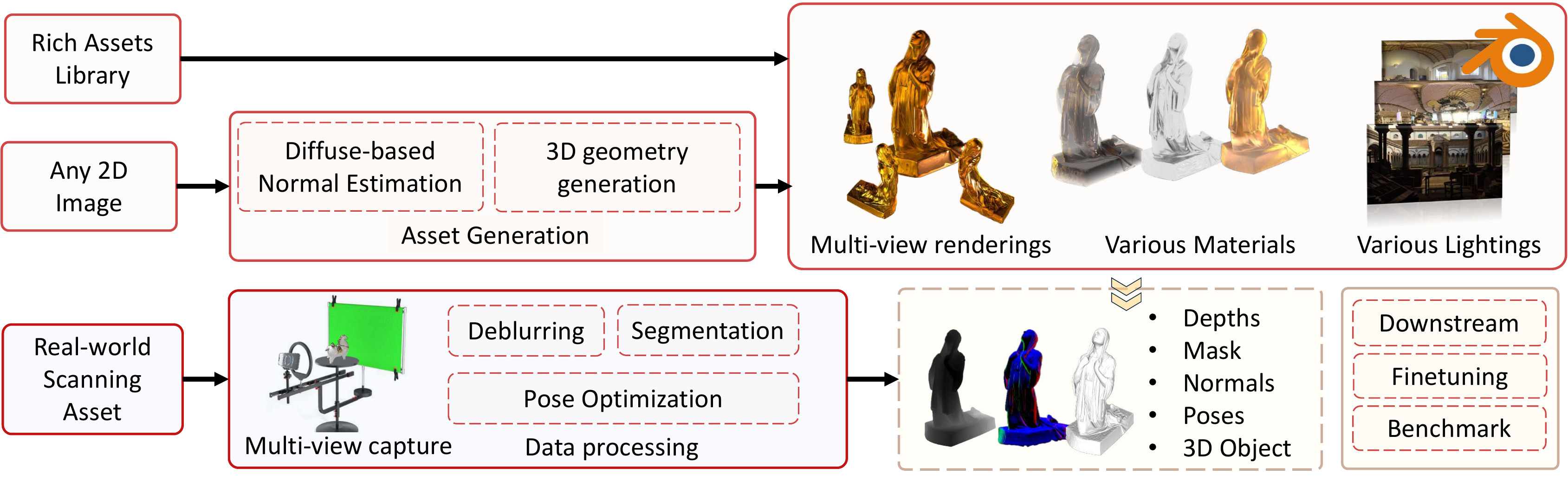}
    \vspace{-3mm}
    \caption{The Dataset Construction and Evaluation Pipeline.}
    \vspace{-5mm}
    \label{fig:pipeline}
\end{figure*}

In this section, we identify the root causes of reconstruction failures for non-Lambertian materials and demonstrate the urgent need for a high-quality dataset of challenging materials for 3D perception tasks.

\subsection{Impact of Material Properties on PBR and Reconstruction Quality}
To investigate how material properties affect physically-based rendering and reconstruction quality, we conducted an extensive experiment analyzing 3D reconstruction algorithm (3DGS~\cite{kerbl20233d}) performance across diverse material parameters. We selected a model and systematically varied its material by spanning four critical parameters: metallic (0 or 1), roughness (0--0.9), IOR (1.0--1.9), and transmission (0 or 1). For each of the 48 material configurations, we trained 3DGS using 50 multi-view images with masked backgrounds and evaluated PSNR on 10 held-out test views.

Figure~\ref{fig:materials_training} illustrates five representative cases showing how different materials interact with light under identical lighting conditions (top row). The reconstructed views (middle row) reveal increasing geometric artifacts---particularly floaters---as light transmission and reflectance violate photometric consistency assumptions. The masked views (bottom row) enable cleaner comparison with ground truth. Figure~\ref{fig:material_parameters} presents the complete parameter sweep across all 48 configurations. The results reveal three distinct failure modes. First, reflective materials with smooth surfaces (roughness=0.0) experience catastrophic failure: metallic materials achieve only ~19~dB PSNR versus ~35~dB for high-roughness non-metallic surfaces---a 45\% degradation. Second, low-roughness surfaces starve correspondence-based methods of texture cues, with PSNR improving by 5~dB as roughness increases from 0.0 to 0.9 in non-metallic materials. Third, transparent materials represent the most critical failure mode, causing a consistent 5.82~dB PSNR drop (19.3\% quality loss) across all configurations. Notably, higher refractive indices progressively worsen performance: transparent materials exhibit PSNR values ranging from 19.9~dB (IOR=1.0) to 27.9~dB (IOR=1.9), confirming that stronger refraction increasingly violates epipolar geometry. For more detailed results and analysis, please refer to Suppls.~\ref{sup-appendix:material_parameter_impact},~\ref{sup-appendix:root cause analysis}.

\subsection{Root Cause: Algorithmic Assumptions vs. Physical Reality}
These failures stem from a fundamental mismatch between algorithmic assumptions and physical light transport. SOTA MVS methods---both classical and learning-based---rely on \textit{photometric consistency}: the assumption of view-invariant, Lambertian-like surfaces. This assumption breaks down catastrophically for non-Lambertian materials. 

Reflective materials violate photometric consistency through view-dependent appearance governed by their BRDF, causing algorithms to misinterpret specular highlights as geometric features. Low-texture surfaces lack the high-frequency features needed for robust correspondence matching, resulting in ambiguous reconstructions. Transparent materials represent the most severe case, breaking both photometric consistency and the geometric assumption of linear light propagation: refraction invalidates the epipolar constraints underpinning multi-view triangulation, causing complete reconstruction breakdown.

\textit{These are not edge cases but systematic failures arising from oversimplified computational models of complex physical phenomena.} Our observation underscores the necessity for reconstruction techniques inherently aware of material-light interactions---and the critical need for datasets that expose these challenges.

\section{Dataset Construction and Statistics}

We begin by comparing \method to existing related datasets and presenting its detailed statistics. We then describe the framework used to construct \method (Figure~\ref{fig:pipeline}). \method comprises two complementary subsets: a synthetic collection and a real‐world scan set. The synthetic data are rendered as photorealistic RGB images in Blender, drawing on two asset sources: a large‐scale existing library and 3D generated models derived from 2D image references. The real‐world subset consists of objects with challenging materials captured under diverse lighting conditions using commodity scanning devices, thereby reflecting common acquisition scenarios.

\subsection{Statistics Overview}
We present a photorealistic, object-centric dataset designed to enhance 3D reconstruction quality across a diverse range of materials, shapes, and lighting conditions. This dataset supports various tasks, including image matching~\cite{sun2021loftr, edstedt2024roma, wang2024efficient}, camera pose estimation~\cite{wang2024dust3r, yang2025fast3r, dong2024reloc3r}, novel-view synthesis~\cite{kerbl20233d,muller_instant_2022}, and a series of other perception tasks (see Appendix for details). \method distinguishes itself from existing datasets by offering a large-scale synthetic-real hybrid collection, specifically curated to address challenging materials and lighting scenarios. Table~\ref{tab:comparison} provides a detailed comparison between \method and existing datasets, while Table~\ref{tab:statistics} outlines the comprehensive statistics characterizing \method.

\begin{figure}[h!]
    \centering
    
    \begin{subfigure}[b]{0.48\linewidth}
        \includegraphics[width=\linewidth]{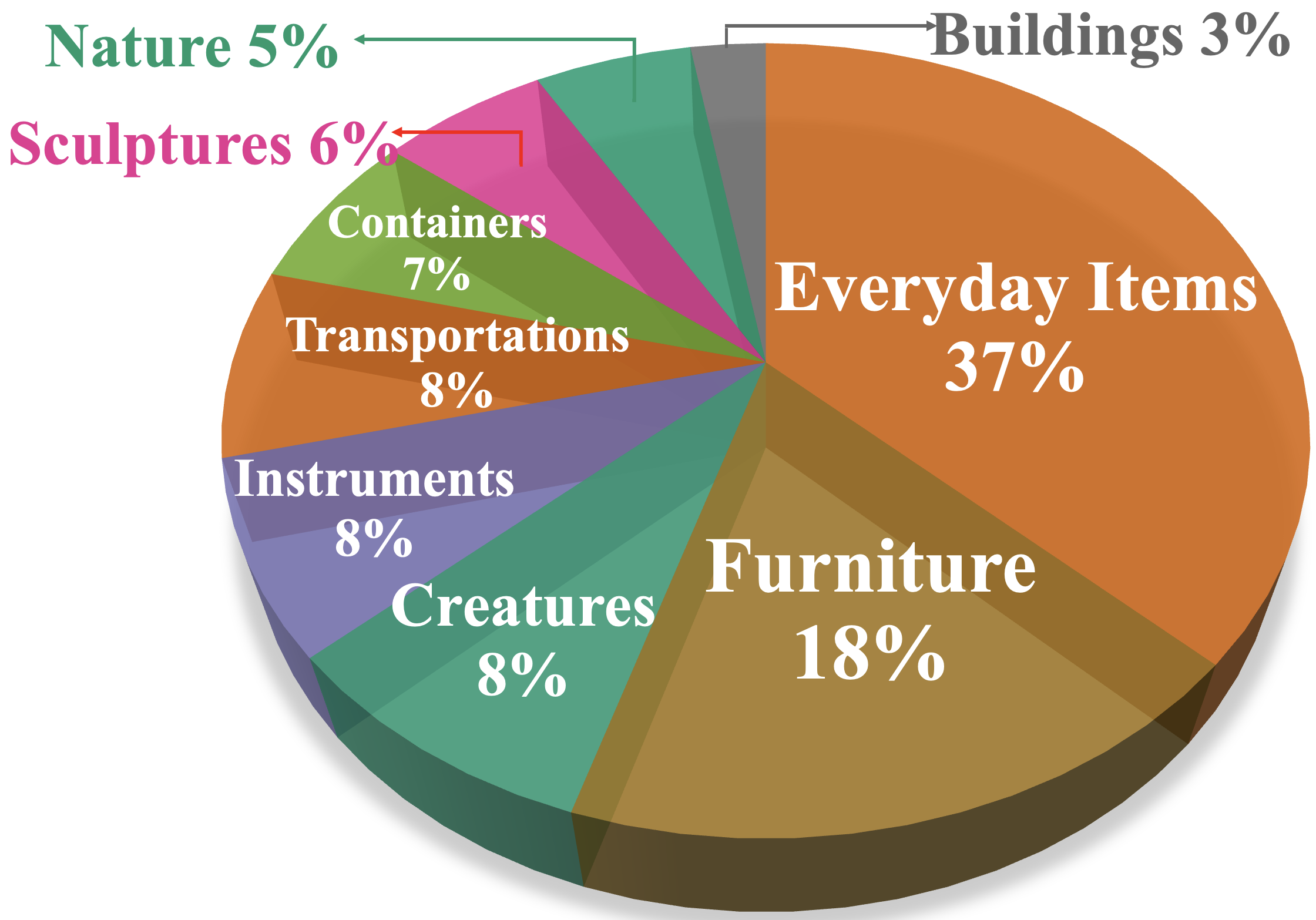}
        \caption{}
        \label{fig:category_count}
    \end{subfigure}
    \hfill
    \begin{subfigure}[b]{0.45\linewidth}
        \includegraphics[width=\linewidth]{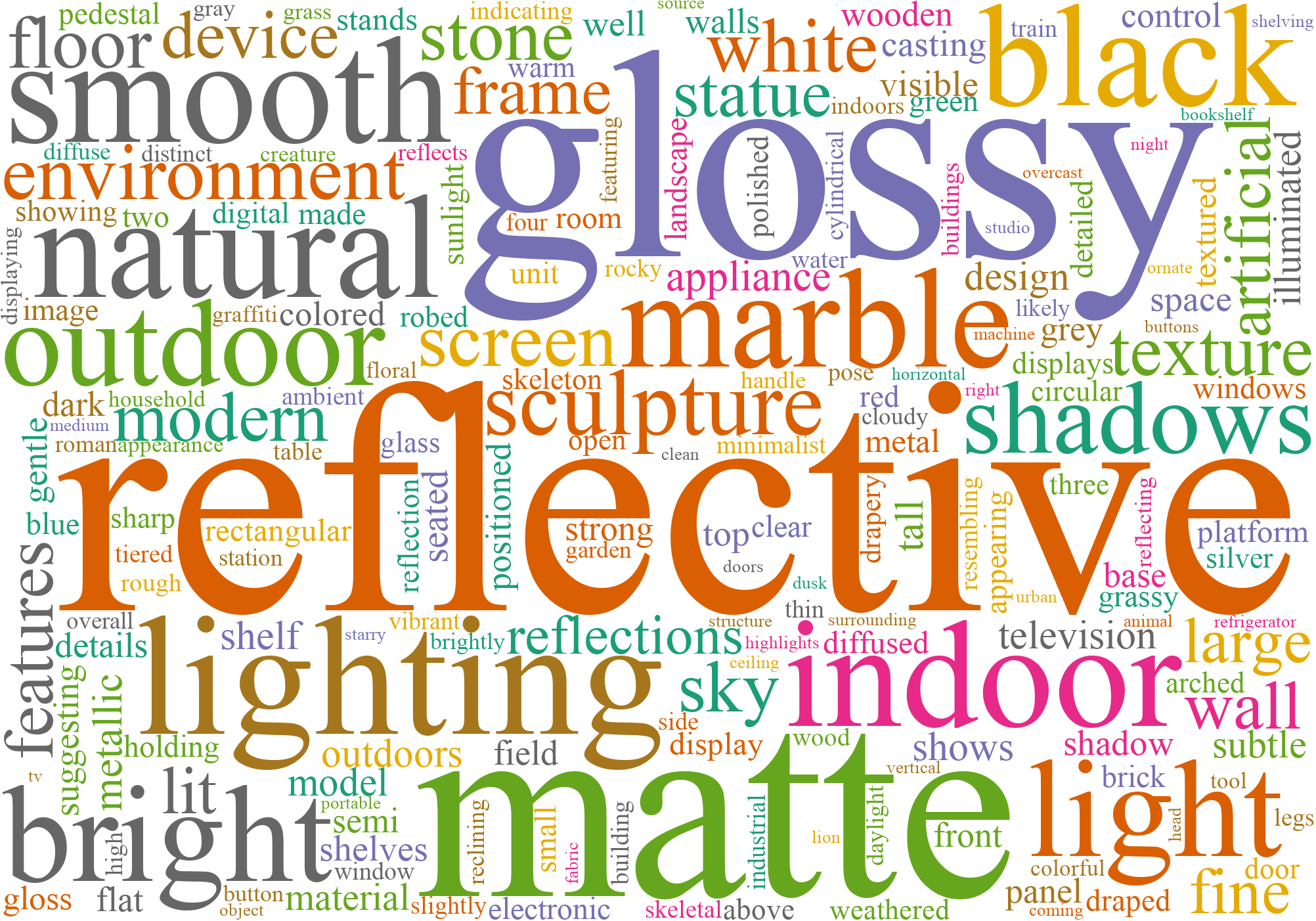}
        \caption{}
        \label{fig:wordcloud}
    \end{subfigure}
    \vspace{-2mm}

    \caption{Overview of synthetic dataset statistics. (a) Distribution of 9 high-level categories. (b) A word cloud visualizing common terms in our LLM-generated textual descriptions}
    \label{fig:dataset_overview}
\end{figure}

Figure~\ref{fig:dataset_overview} provides a comprehensive overview of our synthetic dataset statistics. The dataset is divided into nine high-level categories as shown in Figure~\ref{fig:category_count}. While \textit{Everyday Items} is the largest category, the dataset maintains a balanced distribution across other diverse classes.  Figure~\ref{fig:wordcloud} visualizes our textual descriptions as a word cloud, where terms like \textit{reflective}, \textit{lighting}, and \textit{glossy} confirm our core focus on complex materials and illumination. These detailed labels are not only crucial for analysis but are also designed to support generation-related tasks, such as text-to-3D asset creation, as detailed further in Suppl.~\ref{sup-appendix:generation_annotations}.

\begin{table*}[!t]
\centering
\caption{Comparison between other related datasets. The symbol ``\#'' denotes the total count, ``PBR'' refers to physically-based rendering, and ``w/ Real'' refers to containing real dataset.}
\vspace{4pt}
\label{tab:comparison}
\resizebox{2\columnwidth}{!}{
\begin{tabular}{@{}lcccccccc@{}}
\toprule
Dataset                                    & Type                & Transparent & Reflection & Low-Texture & Relighting & \#Instances & PBR & w/ Real \\ \midrule
MV Reflectance~\cite{oxholm2014multiview}  & Objects             & \xmark      & \cmark     & \xmark      & \xmark     & $<$20         & \cmark                           & \xmark      \\
NeRO~\cite{liu2023nero}                    & Objects             & \xmark      & \cmark     & \cmark      & \cmark     & 8           & \cmark                           & \cmark      \\
ShapeNet-Intrinsics~\cite{shi2017learning} & Objects             & \xmark      & \xmark     & \cmark      & \cmark     & 30K         & \xmark                           & \xmark      \\
DTU~\cite{aanaes2016large}                 & Objects             & \xmark      & Few        & \xmark      & \xmark     & 100+        & -                                & \cmark      \\
Tanks and Temples~\cite{Knapitsch2017}     & Objects/Scenes      & \xmark      & \xmark     & \xmark      & \xmark     & 21          & -                                & \cmark      \\
BlendedMVS~\cite{yao2020blendedmvs}        & Objects/Scenes & \xmark      & \xmark     & \xmark      & \xmark     & 113         & \cmark                           & \xmark      \\
ABO~\cite{collins2022abo}                  & Objects             & \xmark      & \xmark     & \xmark      & \xmark     & 8K          & \cmark                           & \cmark      \\

RTMV~\cite{tremblay2022rtmv}        &    Objects/Scenes         &  \xmark       & Few    & Few    &  \xmark  & $\sim$2000 & \cmark  & \xmark  \\
Objaverse~\cite{deitke2023objaverse}       & Objects                & \xmark      & \xmark     & \cmark      & \xmark     & 800K+       & \xmark                           & \xmark      \\
OpenMaterials~\cite{dang2024openmaterial}                             & Objects             & \xmark      & \cmark     & \cmark      & \cmark     & 1001        & \cmark                           & \xmark      \\
\textbf{\method (Ours)}                    & Objects             & \textcolor{green}{\cmark}      & \textcolor{green}{\cmark}     & \textcolor{green}{\cmark}      & \textcolor{green}{\cmark}     & 120K+       & \textcolor{green}{\cmark}                           & \textcolor{green}{\cmark}      \\ \bottomrule
\end{tabular}
}
\end{table*}

\begin{table}[h!]
\centering
\caption{Summary statistics of the \method dataset. }
\vspace{-3mm}
\label{tab:statistics}
\normalsize
\resizebox{\columnwidth}{!}{
\begin{tabular}{@{}cccccccc@{}}
\toprule
          & \#Shapes & \#Materials & \#Lighting                & \#Instances & \#Views              & \#Frames        \\ \midrule
Synthetic & 12K+     & 22          & \numscenes & 120K+       & 60 & 7M+    \\
Real      & 300+     & $>$50      & 5                         & 1000+       & 100+   & 120K+    \\ \bottomrule
\end{tabular}
}
\vspace{-3mm}
\end{table}

\subsection{Synthetic Data Generation Pipeline}
\label{sec:blender pipeline}

The synthetic component of \method comprises objects drawn from a wide variety of categories, including household items, vehicles, statues, electronics, and other everyday artifacts. It also includes objects with symmetric geometry or low-texture appearances, such as accessories and minerals, to better reflect practical industrial scenarios.

\noindent \textbf{Shapes, Materials and Lighting.} 
We carefully collected over 10K high-quality shapes from scanned object repositories and 3D asset databases covering art, industry, and nature domains. Additionally, we generated more than 2K shapes representing everyday items to ensure diversity. For more details, please refer to \S~\ref{sec:2d_3d}.

In terms of materials, we incorporate \nummaterials types that are both commonly encountered and challenging for accurate 3D reconstruction. These are categorized into five groups: \textit{Diffuse} (e.g., concrete, matte surfaces), \textit{Transparent} (e.g., glass, clear acrylic), \textit{Metallic} (e.g., steel, chrome), \textit{Glossy-Textured} (e.g., polished wood with grain patterns), and \textit{Glossy-Low-Texture} (e.g., ceramic glazes).

\begin{figure*}[t!]
\vspace{-2mm}
    \centering
    \includegraphics[width=\linewidth]{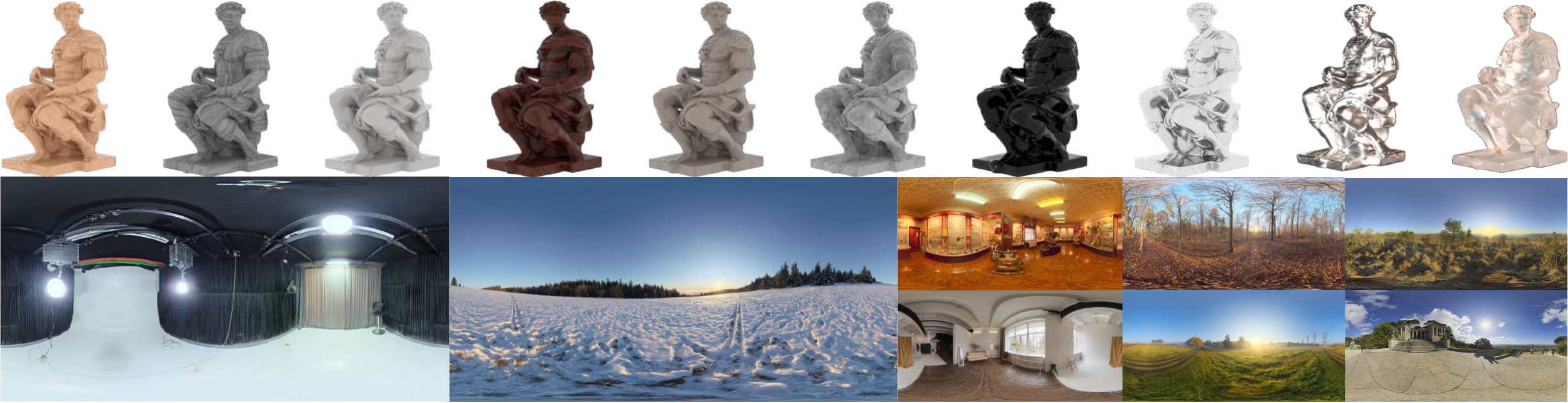}
    \vspace{-4mm}
    \caption{Overview of objects with various materials and lighting conditions in the dataset}
    \vspace{-4mm}
\label{fig:texure and lighting conditions}
    
\end{figure*}

For realistic lighting simulation, we utilize \numscenes HDRI environment maps. These lighting conditions span a wide range of real-world scenarios—indoor and outdoor settings, various times of day, and weather conditions, as well as other nuanced variations—ensuring comprehensive coverage of diverse lighting environments. Figure~\ref{fig:texure and lighting conditions} showcases a subset of the objects, materials, and environment maps used in our dataset. Further, we simulate local illumination by placing 1–2 finite-distance point lights on the upper hemisphere (Suppl.~\ref{sup-appendix:near illumination}). To maximize diversity, each object is selected to pair with different materials and lighting conditions, resulting in over 120K synthetic object instances.

\begin{figure}[t]
    \centering
    \includegraphics[width=\linewidth]{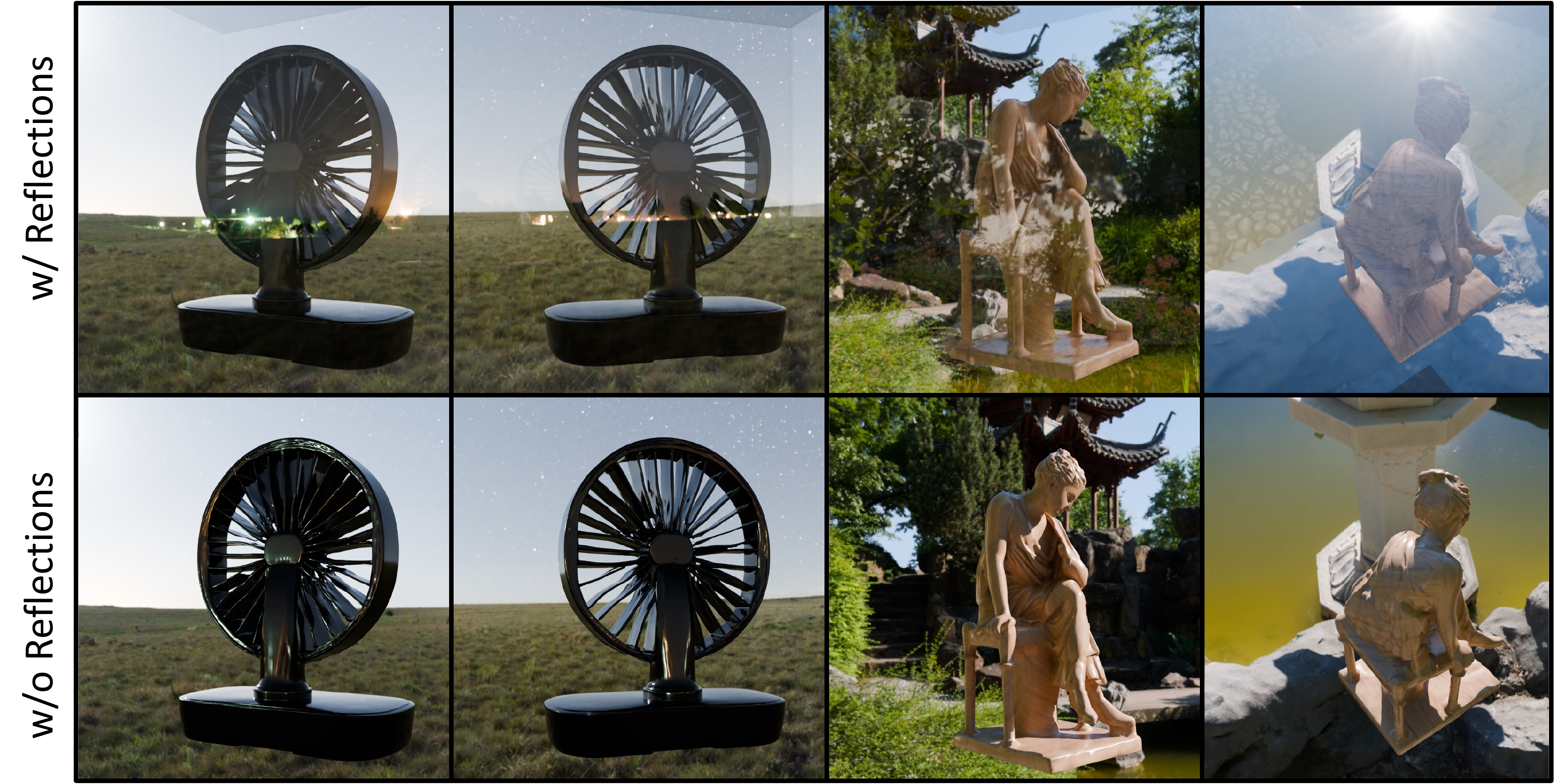}
    \caption{Multi-view Specular Reflection}
    \vspace{-2mm}
    \label{fig:reflection}
\end{figure}

\begin{figure}[t]
    \centering
    \includegraphics[width=\linewidth]{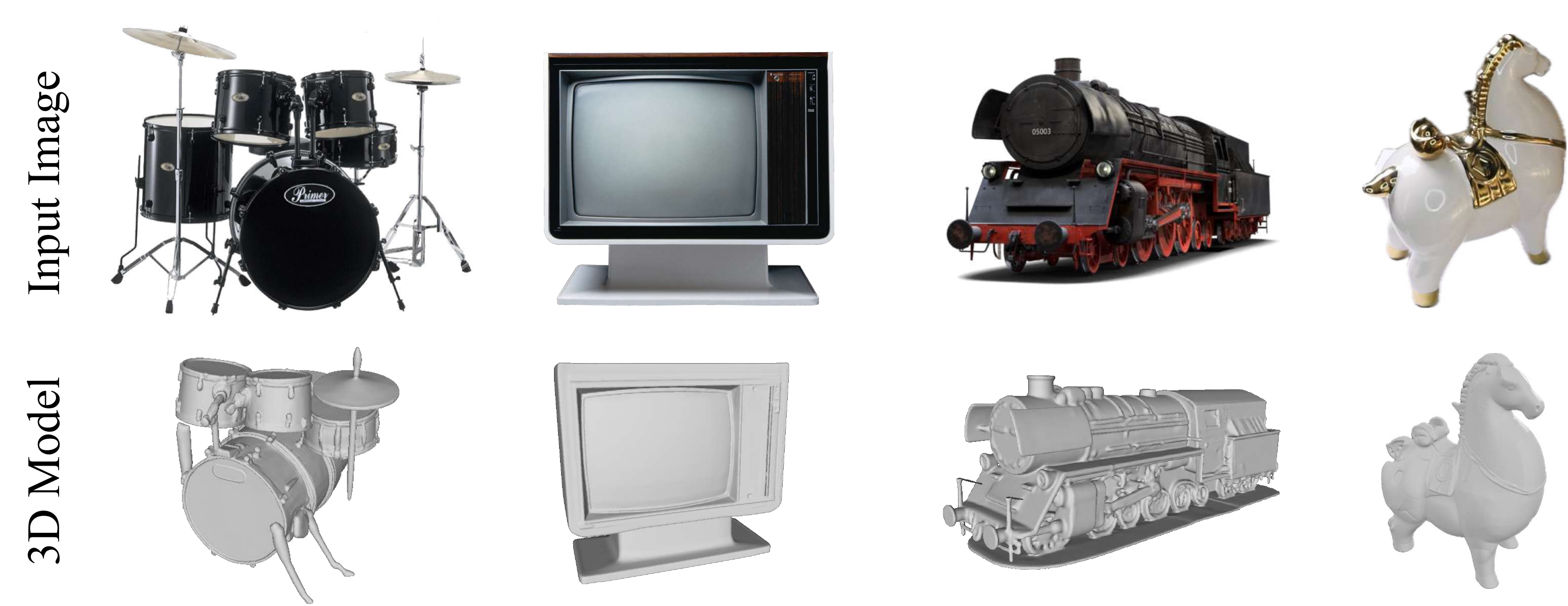}
    \caption{3D Object Generation Given 2D Reference.}
    \vspace{-4mm}
    \label{fig:2d_3d_generation}
\end{figure}

\noindent \textbf{Specular Reflection.} To simulate common specular reflection scenarios, we position a glass sheet between the object and the camera following~\cite{wang2025flash, SIR2pami, zhu2024revisiting}. Unlike previous works~\cite{SIR2-iccv17,wang2025flash,zhu2024revisiting,SIR2pami,lei2021robust} that capture only a single angle under limited lighting conditions, we provide data from 60 different angles under hundreds of lighting conditions to comprehensively simulate reflection effects. In these settings, the glass reflects the surrounding environment, introducing complex optical effects that create significant challenges for 3D perception tasks. The top row of Figure~\ref{fig:reflection} shows multi-view reflections. Detailed explanations of the light behavior are provided in Suppl.~\ref{sup-appendix:light_behavior}.

\noindent \textbf{Rendering and Output.} For each instance, we provide extensive ground truth annotations generated using a physically-based rendering (PBR) engine in Blender. These include 60 multi-view images per instance, high-resolution RGB images rendered at $1000 \times 1000$ resolution, 3D geometry in both point cloud and mesh formats, object segmentation masks, dense depth maps, and surface normal maps. This rich set of annotations supports the detailed evaluation of reconstruction quality from multiple perspectives.

\subsection{Synthesis Data Generation with 2D Reference}
\label{sec:2d_3d}

To enrich shape diversity beyond a predefined library, we developed an efficient pipeline to automatically generate 3D assets from 2D images. We use real-world and LLM-generated 2D images as references to synthesize 3D models via diffusion-based methods~\cite{ye2025hi3dgen, ye2024stablenormal}. This process involves estimating normals and depth, reconstructing a mesh, and refining it to a canonical pose. These models are then rendered in Blender with diverse PBR materials and HDR lighting (\S~\ref{sec:blender pipeline}), providing a lightweight way to scale \method and enhance its realism. Figure~\ref{fig:2d_3d_generation} shows 3D models (bottom row) generated from 2D reference images (top row). The 2D references include three columns of GPT-4o generated images and one column of a real-world capture. Further qualitative results are available in Suppls.~\ref{sup-appendix:3dgen},~\ref{sup-appendix:more qualitative examples}.

\subsection{Real-World Capture}
\vspace{-1mm}

\begin{figure}[t!]
    \centering
    \includegraphics[width=\linewidth]{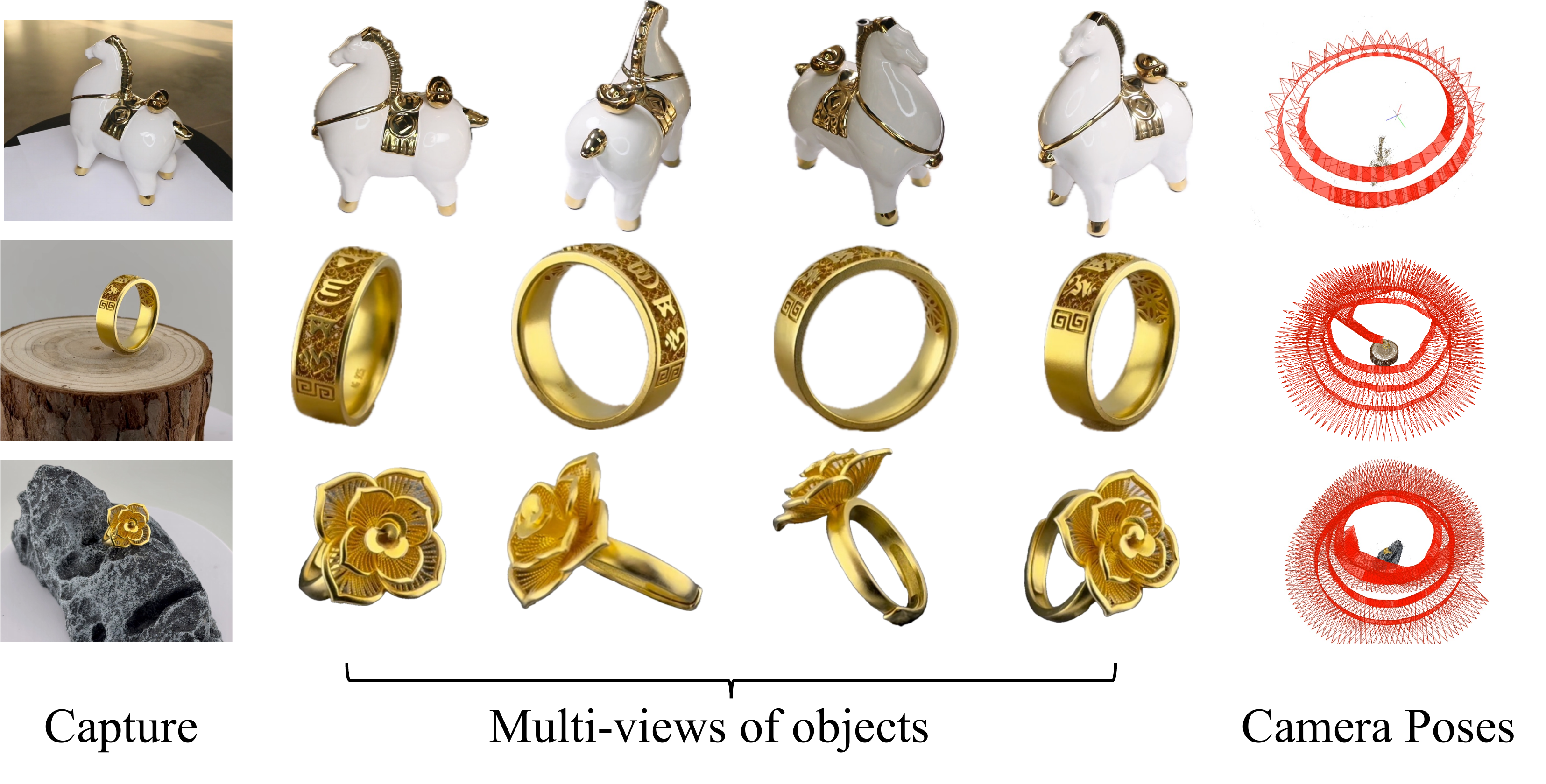}
    \vspace{-5mm}
    \caption{Qualitative examples of capturing reflective objects using rotating platforms.}
    \vspace{-5mm}
    \label{fig:real-data}
\end{figure}

We capture real-world data using an iPhone 16 Pro, recording 1080×1920 video at 30 FPS with default camera settings to simulate typical capture conditions. A primary challenge in this setting is that standard camera pose estimation algorithms fail when objects have challenging materials (e.g., reflective or low-texture) that lack stable, view-invariant features. To circumvent this, our capture protocol is designed to separate the pose estimation task from the object itself. We place the target object on a highly detailed base, which serves as a stable tracking marker. This entire assembly is then placed on a rotating platform (Figure~\ref{fig:pipeline}) to ensure a smooth, stable 360-degree capture path. Our processing pipeline then involves first estimating robust camera poses from the video by tracking the detailed base of the object using RealityScan~\cite{RealityScan2025}. With the camera poses secured, we then employ SAM 2~\cite{ravi2024sam} to segment and remove both the base and the background. This two-stage process yields accurate poses for the challenging objects, and Figure~\ref{fig:real-data} presents qualitative results from our scans. 

\section{Experiments}
\vspace{-1mm}
\subsection{Benchmark}
\vspace{-1mm}

To facilitate standardized evaluation and support research in 3D reconstruction and novel view synthesis, we establish a benchmark based on our proposed \method dataset. The benchmark spans both synthetic and real-world object-centric scenes. For evaluation, 80\% of the data is allocated for training, 10\% for validation, and remaining 10\% for testing. 
Our benchmark supports multiple tasks, including (i) image matching, (ii) structure-from-motion, (iii) novel view synthesis, (iv) reflection and highlight removal, and (v) object relighting.

\subsection{Image Matching}
\vspace{-1mm}

We evaluate image matching performance using three metrics: AUC@5$^\circ$, AUC@10$^\circ$, and AUC@20$^\circ$. These metrics measure the area under the curve of pose accuracy using thresholds $\tau = 5^\circ$, $10^\circ$, and $20^\circ$, based on the minimum of rotation and translation angular errors. We evaluate the methods—covering sparse, semi-dense, and dense image matching strategies—on a selected subset of 1,000 Roman statue instances. The combined results on \method and MegaDepth~\cite{li2018megadepth} are presented in Table~\ref{tab:image_matching}.

Matching objects with reflective, transparent, or low-texture surfaces across different views remains a significant challenge. The results demonstrate that existing methods often struggle to establish accurate correspondences in such cases due to inconsistent appearance, lack of distinctive texture, and view-dependent distortions, and leading to low pose accuracy, while these methods achieve much better results~\cite{edstedt2024roma, wang2024efficient} on benchmarks such as MegaDepth~\cite{li2018megadepth}. This performance gap highlights the importance of developing high-quality datasets that include these challenging yet commonly encountered scenarios.

\begin{figure}[t!]
    \centering
    \includegraphics[width=\linewidth]{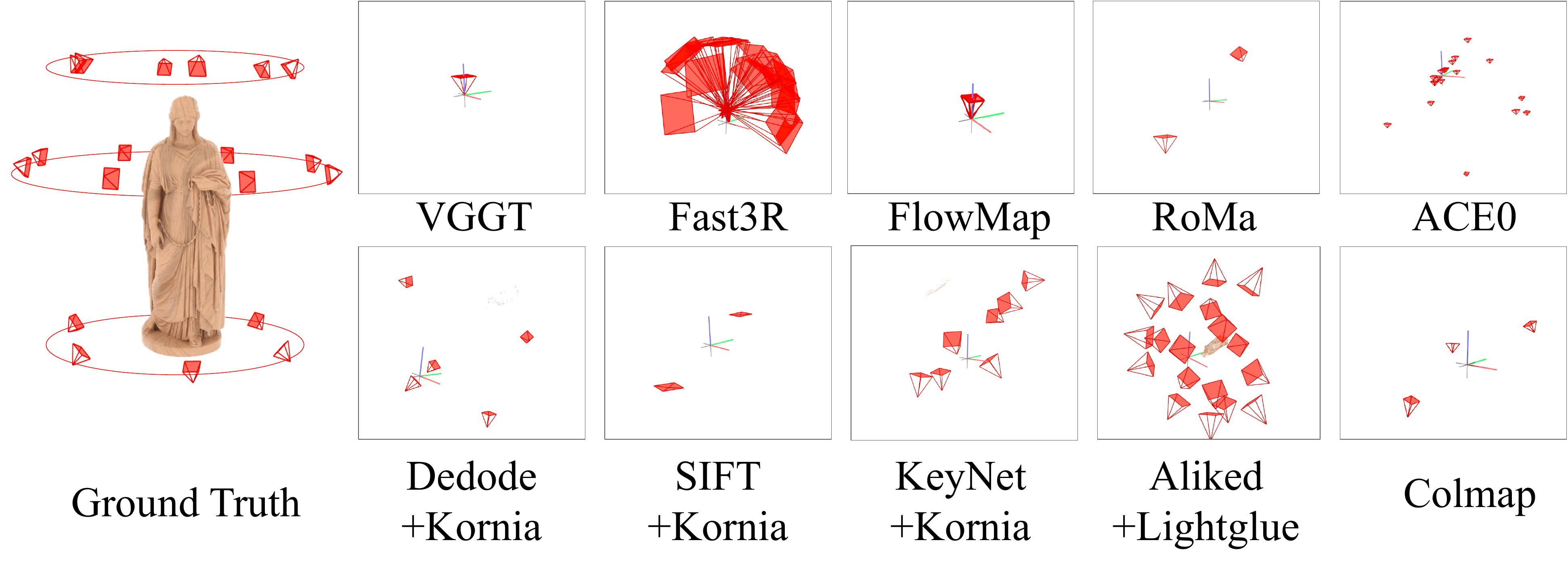}
    \vspace{-6mm}
    \caption{Camera Pose Estimation}
    \label{fig:sfm}
\end{figure}

\begin{table}[t]
\footnotesize
\centering
\caption{Benchmark Image Matching Performance on the \method dataset. Italic numbers represent the results on the MegaDepth dataset~\cite{li2018megadepth}}

\resizebox{\columnwidth}{!}{
\begin{tabular}{@{}lccc@{}}
\toprule
Method      & AUC@5°$\uparrow$        & AUC@10°$\uparrow$       & AUC@20°$\uparrow$       \\ \midrule
SuperPoint~\cite{superpoint} + NN     & 11.2 (\textit{31.7})          & 20.8 (\textit{46.8})          & 27.7 (\textit{60.1})          \\
SuperPoint~\cite{superpoint} + SuperGlue~\cite{sarlin2020superglue}     & 15.2 (\textit{49.7})& 31.0 (\textit{67.1})& 39.9 (\textit{80.6})\\
SuperPoint~\cite{superpoint} + LightGlue~\cite{lindenberger2023lightglue}     & 15.8 (\textit{49.9}) & 31.3 (\textit{67.0})        & 40.1 (\textit{80.1})         \\
LoFTR~\cite{sun2021loftr}               & 19.8 (\textit{52.8})         & 35.6 (\textit{69.2})         & 39.2 (\textit{81.2})         \\
AspanFormer~\cite{chen2022aspanformer} & 20.3 (\textit{55.3})         & 36.9 (\textit{71.5})         & 41.1 (\textit{83.1})          \\
ELoftr~\cite{wang2024efficient}      & 21.3  (\textit{56.4})        & 36.2 (\textit{72.2})            & 41.9 (\textit{83.5})\\
ROMA~\cite{edstedt2024roma}        & 32.1 (\textit{62.6})         & 47.5 (\textit{76.7})         & 59.1 (\textit{86.3})         \\ \bottomrule
\end{tabular}
}
\label{tab:image_matching}

\end{table}

\subsection{Structure from Motion}

SfM aims to reconstruct sparse 3D point clouds while simultaneously estimating camera parameters from a collection of input images.\textit{ Accurate camera pose estimation is essential for downstream multi-view 3D reconstruction tasks, as inaccuracies in estimated poses often lead to geometric inconsistencies and visible artifacts in the reconstructed models, as illustrated in }Figure~\ref{fig:artifact}. In this benchmark, we evaluate 10 representative methods for estimating camera parameters from object-centric, multi-view images in \method. To ensure the evaluation focuses on the object geometry itself, we remove the background while preserving lighting effects on the object surfaces across various materials. This prevents the background from providing auxiliary features that may bias pose estimation, forcing the methods to rely solely on object-intrinsic features to compute relative poses between image pairs. Figure~\ref{fig:sfm} reveals that most existing methods struggle to recover accurate camera parameters under these conditions.


\begin{table}[t!]
\footnotesize
\caption{Benchmark NVS Performance on the \method dataset across material categories, measured by PSNR$\uparrow$.}
\setlength{\tabcolsep}{5pt}
\resizebox{\columnwidth}{!}{
\begin{tabular}{@{}lccccc@{}}
\toprule
\textbf{Method} & \textbf{Diffuse} & \textbf{Transparent} & \textbf{Metallic} & 
\textbf{\begin{tabular}[c]{@{}c@{}}Glossy-\\Textured\end{tabular}} & 
\textbf{\begin{tabular}[c]{@{}c@{}}Glossy-\\Low-Texture\end{tabular}} \\ 
\midrule
Instant-NGP~\cite{mueller2022instant} & 36.12 & 19.20 & 25.59 & 34.01 & 26.52 \\
3DGS~\cite{kerbl20233d}               & 36.99 & 20.20 & 27.02 & 34.10 & 27.62 \\
Splatfacto~\cite{tancik2023nerfstudio} & 37.32 & 21.31 & 28.61 & 34.21 & 28.01 \\
2DGS~\cite{huang20242d}               & 36.77 & 17.12 & 28.46 & 34.42 & 27.97 \\ \bottomrule
\end{tabular}
}
\label{tab:nvs}
\end{table}

\begin{table}[t]
\centering
\caption{Benchmark Surface Reconstruction Performance on the \method dataset across material categories, measured by Chamfer Distance $\downarrow$.}
\label{tab:surface_recon}
\resizebox{\columnwidth}{!}{
\begin{tabular}{lccccc}
\toprule
\textbf{Method} & \textbf{Diffuse} & \textbf{Transparent} & \textbf{Metallic} & 
\textbf{\begin{tabular}[c]{@{}c@{}}Glossy-\\Textured\end{tabular}} & 
\textbf{\begin{tabular}[c]{@{}c@{}}Glossy-\\Low-Texture\end{tabular}} \\
\midrule
Instant-NGP~\cite{mueller2022instant}    & 0.087 & 0.185 & 0.175 & 0.132 & 0.149 \\
Neus2~\cite{wang2023neus2}               & 0.079 & 0.165 & 0.158 & 0.105 & 0.121 \\
2DGS~\cite{huang20242d}                  & 0.060 & 0.142 & 0.121 & 0.086 & 0.098 \\
PGSR~\cite{chen2025pgsr}                 & 0.062 & 0.502 & 0.412 & 0.162 & 0.228 \\
\bottomrule
\end{tabular}
}
\end{table}

\subsection{NVS and Surface Reconstruction}

We evaluate six representative methods on novel view synthesis and surface reconstruction across five material categories, each containing five objects of varying geometry: \textit{Diffuse} materials exhibit Lambertian reflectance without specular highlights; \textit{Transparent} materials feature strong light transmission and refraction; \textit{Metallic} materials display high specularity and colored reflections; \textit{Glossy-Textured} materials combine specular reflections with surface details; and \textit{Glossy-Low-Texture} presents low-texture specular surfaces.

\noindent \textbf{Novel View Synthesis.} As shown in Table~\ref{tab:nvs}, all evaluated methods excel on diffuse materials, achieving PSNR scores above 36 dB. This strong performance is expected, as these methods rely on multi-view photometric consistency, an assumption that diffuse surfaces closely satisfy. In contrast, metallic and glossy-low-texture materials with strong specular reflections violate the view-consistency assumption, resulting in degraded performance. The most significant challenges arise with transparent materials, where all methods struggle severely (PSNR $\sim$17--21 dB), as complex light transmission, refraction, and caustics fundamentally break the color consistency principle that these methods rely upon. Notably, glossy-textured materials achieve relatively high PSNR ($\sim$34 dB) due to the presence of diffuse texture patterns, yet still exhibit artifacts in specular regions. These performance trends are consistent with the qualitative examples shown in Figure~\ref{fig:materials_training}.

\begin{figure}[t!]
    \centering
    \includegraphics[width=1.05\linewidth]{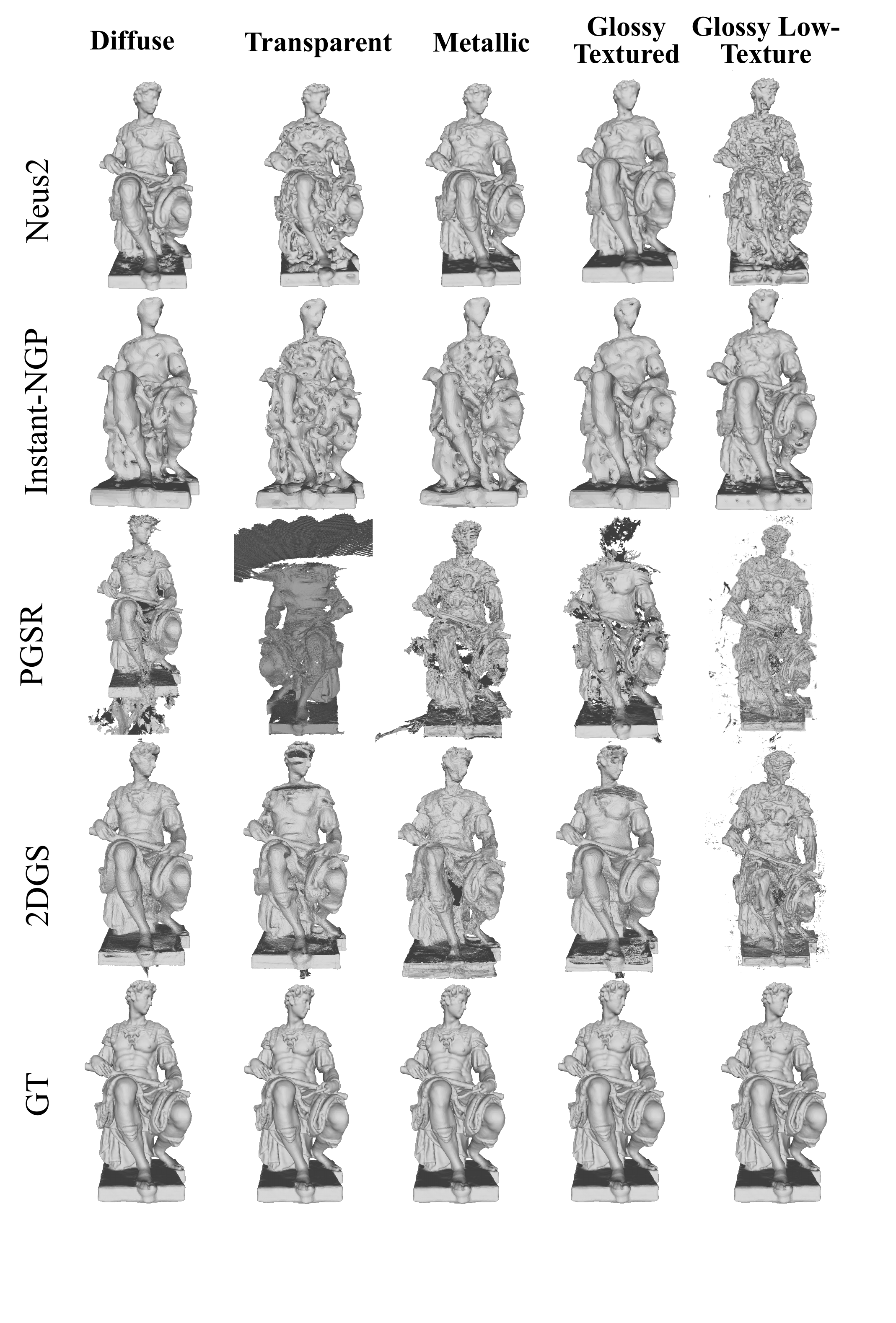}
    \vspace{-10mm}
    \caption{Representative qualitative results of surface reconstruction across various materials.}
    \label{fig:surface}
\end{figure}

\noindent \textbf{Surface Reconstruction.} Table~\ref{tab:surface_recon} presents the surface reconstruction quality across material categories. All methods perform well on diffuse materials, where geometric features can be reliably triangulated from multi-view correspondences. However, reconstruction quality degrades significantly for materials with non-Lambertian appearance. Figure~\ref{fig:surface} demonstrates representative qualitative results.

These results reveal fundamental limitations of current NVS and surface reconstruction methods when handling materials with complex light interactions. Methods relying solely on photometric consistency fail to generalize across the diverse material categories in \method, particularly for reflective, transparent, and low-texture surfaces. 


\subsection{Evaluations on Reflection Removal and Relighting Tasks and Real-World Data}
Due to page constraints, we defer the experimental results for the reflection removal and relighting tasks to Suppls.~\ref{sup-appendix:reflection_removal},~\ref{sup-appendix:relighting}. The evaluation on our real-world dataset is also presented in Suppls.~\ref{sup-appendix:image matching},~\ref{sup-appendix:nvs realworld}.

SOTA methods for reflection removal, relighting, and real-world tasks perform poorly on \method, with results comparable to other challenging real-world datasets. This validates \method as a physically-realistic benchmark. This consistent failure on both our synthetic and real-world instances confirms that SOTA methods, designed for simple Lambertian surfaces, are not robust to the complex, non-Lambertian challenges our dataset exposes.

\section{Discussions}
\textbf{3D Vision and Generation Extensions.} Although this paper emphasizes a subset of 3D vision tasks, \method also supports related problems such as highlight removal, inverse rendering, depth estimation, and surface normal prediction, by providing ground truth labels (Suppl.~\ref{sup-appendix:tasks}).

While our primary focus is on perception tasks, \method's rich annotations open avenues for future research in generative 3D vision, including image-editing~\cite{nam2024contrastive, shuai2024survey, li2024zone}, text-to-3D~\cite{poole2022dreamfusion, yu2023text, ye2024dreamreward, liang2024luciddreamer}, text-to-texture~\cite{richardson2023texture, cao2023texfusion, chen2023text2tex}, image-to-3D~\cite{ye2025hi3dgen, liu2024one, xiang2025structured} and image-to-texture~\cite{hunyuan3d2025hunyuan3d, zeng2024paint3d} in \textit{complex materials and lighting}. To enable such extensions, we annotate each instance with detailed textual descriptions and tags derived from Qwen3-VL-30B-A3B-Instruct~\cite{qwen3technicalreport}. This provides a foundation for integrating \method into downstream generation pipelines. The detailed annotation process for generative 3D vision tasks refer to Suppl.~\ref{sup-appendix:generation_annotations}.

\noindent \textbf{Toward Robust 3D Understanding.} Our findings reveal that existing methods struggle with non-Lambertian and low-texture materials, underscoring the need for new approaches that combine geometric reasoning with photometric modeling. By surfacing these limitations systematically, \method lays the groundwork for developing more generalizable, physically-aware 3D vision systems.

\section{Conclusion}
We presented \method, a large-scale, hybrid dataset designed to advance 3D reconstruction under challenging material conditions—namely, reflectivity, transparency, and low texture. Unlike existing benchmarks, 3DReflecNet combines over 12,000 physically-based synthetic objects with more than 1,000 real-world scans, offering extensive diversity in materials, shapes, and lighting. We further include novel assets generated via diffusion-based shape synthesis and provide comprehensive multi-view annotations to support tasks ranging from image matching, SFM, and novel view synthesis to reflection removal and relighting.

Extensive benchmarking reveals that although state-of-the-art methods perform well on diffuse objects, their performance degrades substantially in the presence of complex optical phenomena. This highlights the need for datasets and models that explicitly address such cases. By releasing \method along with baseline implementations and evaluation tools, we aim to foster research on generalizable and robust 3D vision methods under real-world conditions.

{\small
\vspace{-2mm}
\paragraph{Acknowledgments.}
The work was supported in part by the Guangdong S\&T Programme (Grant No. 2024B0101030002), the Basic Research Project No. HZQB-KCZYZ-2021067 of Hetao Shenzhen-HK S\&T Cooperation Zone, the National Key Research and Development Program of China (Grant No. 2024YFB2907000), the National Natural Science Foundation of China (Grant No. 62293482 and Grant No. 62471423), the Shenzhen Science and Technology Program (Grant No. JCYJ20241202124021028 and Grant No. JCYJ20230807114204010), the Guangdong Talents Program (Grant No. 2024TQ08X346), the Shenzhen Outstanding Talents Training Fund 202002, the Guangdong Provincial Key Laboratory of Future Networks of Intelligence (Grant No. 2022B1212010001) and the Shenzhen Key Laboratory of Big Data and Artificial Intelligence (Grant No. SYSPG20241211173853027).
}

{
    \small
    \bibliographystyle{ieeenat_fullname}
    \bibliography{main/reference.bib}
}

\ifarxiv
    
\ifarxiv 
\newpage
\quad
\newpage

\appendix
\renewcommand{\thesection}{\Alph{section}} 

\startcontents[sections]
\printcontents[sections]{l}{1}{\setcounter{tocdepth}{3}}
\newpage

\fi


\section{Instance Breakdown}
\label{sup-appendix:tasks}

To support various downstream tasks, we provide different types of ground truth data to serve as supervised labels. For each synthesis instance, we provide:
\begin{itemize}
    \item 50 views RGB PNG images of $1,000\times1,000$ resolution
    \item corresponding depth images in exr format 
    \item corresponding normals images in exr format for high precision training
    \item corresponding mask images
    \item point cloud file
    \item Camera internal/external parameters
    \item Physically-based rendering file (in format of .blend).
\end{itemize}

\noindent For each real-world instance, we provide:
\begin{itemize}
    \item $>$60 views RGB PNG images of $1920\times1080$ (or $3840\times2160$) resolution
    \item corresponding mask images
    \item point cloud file
    \item Camera internal/external parameters
\end{itemize}

The design of the instance breakdown can serve different downstream tasks as summarized in Table~\ref{tab:tasks}.

\begin{table*}[h!]
\centering
\caption{Overview of dataset composition and supported downstream applications.}
\begin{tabular}{@{}cc@{}}
\toprule
Design                 & Task                                                  \\ \midrule
Normals                & Normal Estimation~\cite{ye2024stablenormal, ye2025hi3dgen};  Surface Reconstruction~\cite{oechsle2021unisurf, kerbl20233d}            \\
Depths                 & Depths Estimation~\cite{ye2023constraining}                                     \\
Masks                  & Segmentation~\cite{ravi2024sam}, Detection~\cite{superpoint}                                          \\
(non-)Uniform Lighting & Inverse rendering~\cite{liu2023nero},  Highlight Reflection Removal~\cite{Fu_2023_ICCV, Dong_2021_ICCV}      \\
Multi-view Renderings  & Image Matching~\cite{edstedt2024roma, loftr}, Structure from Motion~\cite{incremental}, 3D Reconstruction~\cite{kerbl20233d} \\
w/ (w/o) Glass Reflection  & Specular Reflection Removal~\cite{wang2025flash}                           \\ \bottomrule
\end{tabular}
\label{tab:tasks}
\end{table*}

\section{Near-Field Illumination}
\label{sup-appendix:near illumination}
While standard HDRI environment maps assume illumination from infinity, they could fail to capture the spatial variations of indoor lighting. To enhance realism, we augment the global environment map with near-field lighting, explicitly simulating local effects by positioning one or two finite-distance point lights on the upper hemisphere. (Figure~\ref{fig:local illumination}).

\begin{figure}
\vspace{-1mm}
    \centering
    \includegraphics[width=\linewidth]{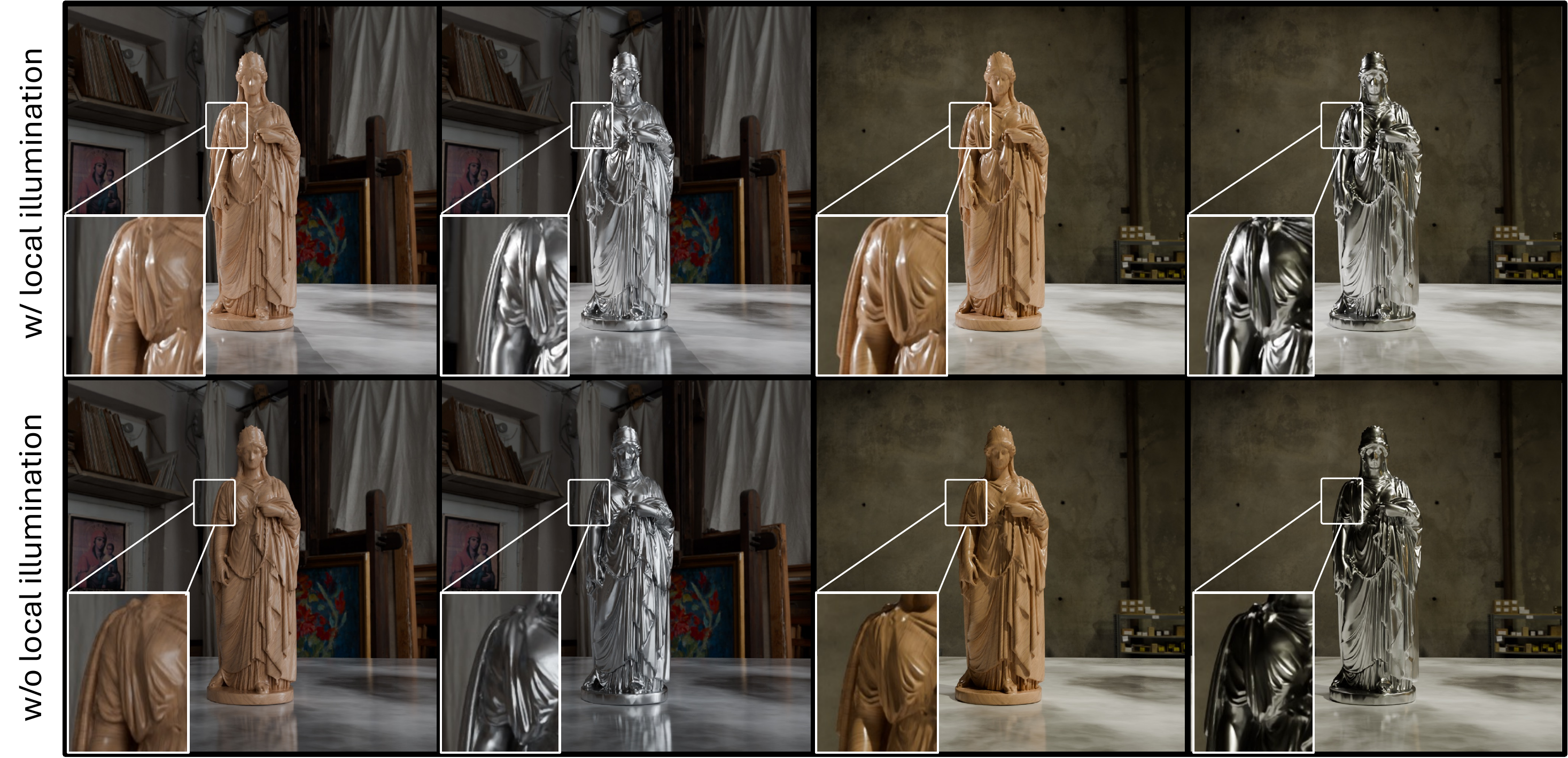}
    \vspace{-5mm}
    \caption{\textbf{Synthetic Enhancement for Indoor Scenes.} Top: Local illumination with finite-distance lights. Bottom: Standard infinite-distance HDRI.}
    \label{fig:local illumination}
    \vspace{-6mm}
\end{figure}

\section{Assets Generation using 2D image}
\label{sup-appendix:3dgen}

\begin{figure*}[h!]
    \centering
    \includegraphics[width=\linewidth]{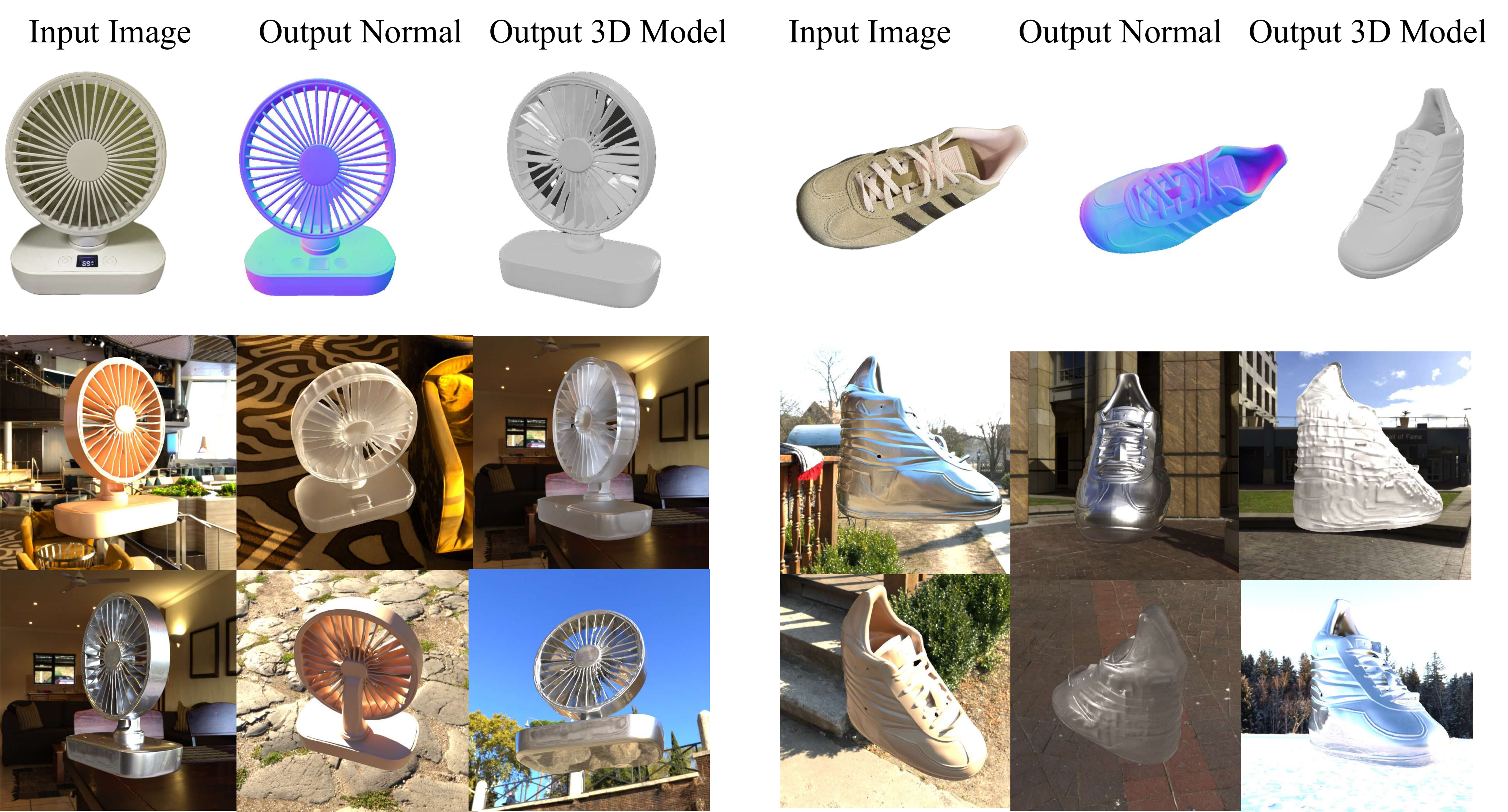}
    \caption{Qualitative Results of 3D Generated Models with Various Materials and Environment Maps. This figure showcases generated models with diverse materials under different lighting. The instance on the left demonstrates a high-quality shape, while the instance on the right shows a failure case that was filtered from our final dataset.}
    \label{fig:3dgen_small}
\end{figure*}

We generated 3D objects from a dataset of over 5K high-quality real-world captures and synthetic images, using either Hunyuan3D-2.1~\cite{hunyuan3d2025hunyuan3d} or Stable3DGen~\cite{ye2025hi3dgen}. The generation process used 50 inference steps and an octree resolution of 320. The resulting objects have an average size of $\sim 52$ Mb and an average of $\sim 25K$ vertices. 

We then performed a manual quality check on the generated assets. While many models, like the electric fan in Figure~\ref{fig:3dgen_small} (left), accurately reproduce the source object's structure, some fail to represent the intended shape faithfully, such as the shoe example in Figure~\ref{fig:3dgen_small} (right). We filtered out these suboptimal results, retaining a final dataset of over 20K high-quality shapes. Figure~\ref{fig:3dgen_shape_material} shows more examples of these selected shapes and materials. Finally, Figure~\ref{fig:3dgen_lighting} showcases the realistic light reflection behavior on a generated steel asset under various lighting conditions.

\section{Image Matching}
\label{sup-appendix:image matching}
We provide additional details on the image matching task. Current methods often struggle to deliver satisfactory image matching accuracy for reflective, transparent, or low-texture objects under varying viewpoints, due to their complex and view-dependent appearance characteristics. Figure~\ref{fig:image_matching_more} show cases Eloftr~\cite{wang2024efficient} performance under these challenging cases. We also provide quantitative results on our real-world captures. For this, we sampled 100 instances to benchmark the performance of several SOTA image matching methods, with results presented in Table~\ref{tab:image_matching_performance}. 

\begin{figure*}[ht!]
    \centering
    \includegraphics[width=\linewidth]{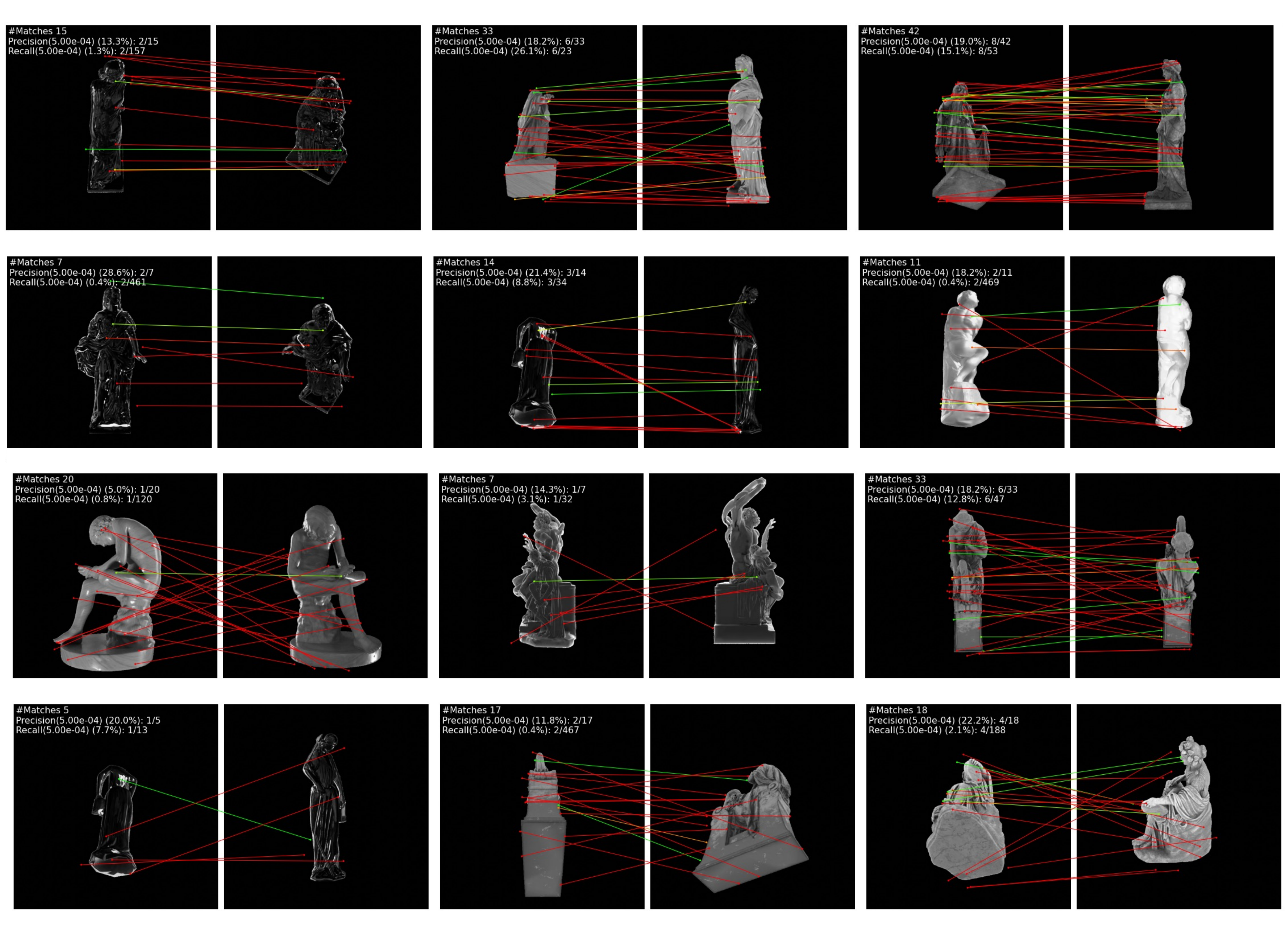}
    \caption{Qualitative results on Efficient Loftr~\cite{wang2024efficient} on reflective, transparent Materials.}
    \label{fig:image_matching_more}
\end{figure*}

\begin{table}[h]
  \centering
  \caption{Evaluation of Image Matching on Real-World Capture}
  \resizebox{\columnwidth}{!}{
  \begin{tabular}{lccc}
    \toprule
    \textbf{Method} & \textbf{AUC@5°$\uparrow$} & \textbf{AUC@10°$\uparrow$} & \textbf{AUC@20°$\uparrow$} \\
    \midrule
    SuperPoint+ NN         & 23.9 & 33.8 & 41.2 \\
    SuperPoint + SuperGlue & 26.2 & 43.1 & 51.3 \\
    SuperPoint+ LightGlue  & 25.8 & 43.5 & 51.9 \\
    LoFTR                  & 27.1 & 45.9 & 50.8 \\
    AspanFormer            & 28.3 & 47.2 & 51.2 \\
    ELoFtr                 & 28.5 & 47.1 & 53.7 \\
    ROMA                   & 34.3 & 49.9 & 59.5 \\
    \bottomrule
  \end{tabular}
  }
  \label{tab:image_matching_performance}
\end{table}

The overall modest scores shown in Table~\ref{tab:image_matching_performance}, with the top-performing method only achieving 59.5 at AUC@20°, indicate that robustly matching our real-world instances of challenging materials remains a significant and unsolved problem.

\section{Detailed Surface Reconstruction Results}

We provide a detailed quantitative comparison of surface reconstruction baselines in Table~\ref{tab:appendix_surface_recon}. The results are broken down by our five main material categories to show how performance varies with material complexity.

\begin{table*}[]
\centering
\caption{Detailed quantitative comparison of surface reconstruction methods on the \method dataset, broken down by material category. `NGP' refers to Instant-NGP~\cite{muller2022instant}}
\resizebox{2\columnwidth}{!}{
\begin{tabular}{lccccccccccccccccccccc}
\toprule
 & \multicolumn{4}{c}{\textbf{Diffuse}} 
 & \multicolumn{4}{c}{\textbf{Transparent}} 
 & \multicolumn{4}{c}{\textbf{Metallic}} 
 & \multicolumn{4}{c}{\textbf{Glossy-Textured}} 
 & \multicolumn{4}{c}{\textbf{Glossy-Low-Texture}} \\
\cmidrule(lr){2-5} \cmidrule(lr){6-9} \cmidrule(lr){10-13} \cmidrule(lr){14-17} \cmidrule(lr){18-21}
\textbf{Metric} 
 & 2DGS & NGP & Neus2 & PGSR 
 & 2DGS & NGP & Neus2 & PGSR 
 & 2DGS & NGP & Neus2 & PGSR 
 & 2DGS & NGP & Neus2 & PGSR 
 & 2DGS & NGP & Neus2 & PGSR \\
\midrule
Hausdorff $\downarrow$ 
 & 0.303 & 1.119 & 0.872 & 0.483 
 & 2.347 & 3.214 & 2.982 & 6.842 
 & 1.862 & 2.873 & 2.641 & 5.923 
 & 1.053 & 1.924 & 1.632 & 4.102 
 & 1.248 & 2.105 & 1.824 & 4.876 \\
PSNR $\uparrow$        
 & 35.32 & 32.96 & 33.55 & 34.06 
 & 22.76 & 20.04 & 21.65 & 17.95 
 & 26.41 & 23.58 & 24.22 & 19.84 
 & 31.05 & 28.10 & 29.32 & 24.32 
 & 29.87 & 26.84 & 27.91 & 22.10 \\
SSIM $\uparrow$        
 & 0.960 & 0.941 & 0.949 & 0.954 
 & 0.811 & 0.762 & 0.789 & 0.742 
 & 0.867 & 0.821 & 0.834 & 0.772 
 & 0.928 & 0.885 & 0.902 & 0.821 
 & 0.912 & 0.863 & 0.884 & 0.795 \\
LPIPS $\downarrow$     
 & 0.070 & 0.098 & 0.087 & 0.088 
 & 0.190 & 0.253 & 0.218 & 0.347 
 & 0.188 & 0.242 & 0.212 & 0.328 
 & 0.121 & 0.179 & 0.148 & 0.256 
 & 0.138 & 0.192 & 0.167 & 0.284 \\
\bottomrule
\end{tabular}
}
\label{tab:appendix_surface_recon}
\end{table*}

All methods, including 2DGS, exhibit a clear performance degradation as material complexity increases. The best results are achieved on Diffuse materials, followed by a noticeable drop on Glossy surfaces, and a severe drop on Metallic and Transparent materials. The failure of methods like PGSR and Instant-NGP is particularly evident in the Hausdorff distance, which explodes on non-Lambertian materials, indicating a catastrophic failure to reconstruct large parts of the geometry.

\section{Highlight and Specular Reflection Removal}
\label{sup-appendix:reflection_removal}

\subsection{Preliminary}

Highlight removal and specular reflection removal address different types of image artifacts caused by light interactions with surfaces and transparent media. Both problems can be formally described using simplified physical models.

Highlight removal focuses on separating the observed image intensity $I$ into its diffuse and specular components:
\begin{equation}
I = I_d + I_s
\end{equation}
where $I_d$ denotes the diffuse component and $I_s$ the specular component. The diffuse component originates from subsurface scattering and re-emission of light, while the specular component arises from direct reflection of incident light. The magnitude and spatial distribution of $I_s$ depend on surface properties such as roughness and material type.

In contrast, specular reflection removal addresses images captured through transparent media, such as glass. In such cases, the observed image $I$ is a mixture of a transmission layer $I_t$ and a reflection layer $I_r$:
\begin{equation}
I = I_t + I_r
\end{equation}

Here, $I_t$ corresponds to the scene visible through the transparent surface, often degraded by refraction and absorption, while $I_r$ results from light reflected off the surface of the medium. These components are often modeled as:
\begin{equation}
I_t = \alpha I_T
\end{equation}
\begin{equation}
I_r = \beta (I_R * k)
\end{equation}
where $I_T$ and $I_R$ denote the original transmission and reflection images, respectively; $\alpha$ and $\beta$ are weighting coefficients; and $k$ is a degradation kernel accounting for blurring or distortion introduced by the reflective surface.

While both tasks involve decomposing a mixed image into multiple layers, they differ in their physical assumptions and application contexts. Highlight removal targets localized reflections on opaque surfaces, whereas specular reflection removal addresses global reflections through transparent materials. Removing specular highlights and reflections enhances image‑matching accuracy and, in turn, improves the final 3D reconstruction.

\subsection{Evaluation of Reflection Removal on \method}

We evaluated four state-of-the-art reflection removal baselines on the \method dataset. For this experiment, we uniformly sampled 1,000 images, applying each method and reporting the average PSNR and SSIM against the ground truth transmission layer. The quantitative results are presented in Table~\ref{tab:reflection_removal}. 

\begin{table}[htbp]
\centering
\caption{Quantitative comparison of reflection removal baselines on 1,000 images from the \method dataset. Higher is better for both metrics.}
\label{tab:reflection_removal}
\resizebox{\columnwidth}{!}{
\begin{tabular}{@{}lcccc@{}}
\toprule
\textbf{Metrics} & ERRNet~\cite{wei2019single} & DSRNet~\cite{hu2023single} & RRW~\cite{zhu2024revisiting} & DSIT~\cite{hu2024single} \\
\midrule
PSNR ($\uparrow$) & 19.49 & 22.92 & 23.11 & 24.07 \\
SSIM ($\uparrow$) & 0.701 & 0.782 & 0.791 & 0.795 \\
\bottomrule
\end{tabular}
}
\end{table}

\noindent The results in Table~\ref{tab:reflection_removal} show a clear performance trend, with DSIT achieving the best results (24.07 PSNR / 0.795 SSIM), followed by RRW and DSRNet. These modest scores are comparable to those reported on other challenging real-world datasets. This consistency validates that our experimental setup is effective and that \method serves as a challenging and physically-realistic benchmark.

\section{Evaluation of NVS and Surface Reconstruction on Real-World Captures}
\label{sup-appendix:nvs realworld}
To validate the challenges of our dataset, we benchmarked SOTA methods on our real-world captures. The following tables present the performance for Novel View Synthesis (NVS) and Surface Reconstruction tasks.

\begin{table}[htbp]
  \centering
  \caption{Novel View Synthesis performance on the real-world dataset. Metrics: PSNR$\uparrow$ (LPIPS$\downarrow$)}
  \begin{tabular}{@{}lc@{}}
    \toprule
    \textbf{Method} & \textbf{Real-world Dataset} \\
    \midrule
    Instant-NGP~\cite{muller2022instant}     & 28.12 (0.025) \\
    3DGS~\cite{kerbl20233d}            & 30.99 (0.020) \\
    Splatfacto~\cite{tancik2023nerfstudio}      & 32.07 (0.020) \\
    2DGS~\cite{huang20242d}            & 33.16 (0.021) \\
    \bottomrule
  \end{tabular}
  \label{tab:nvs_performance_real}
\end{table}

\noindent \textbf{NVS Analysis.}
The NVS results on our real-world data in Table~\ref{tab:nvs_performance_real} show a clear performance hierarchy, with 2DGS (33.16 PSNR) performing best, followed by Splatfacto (32.07 PSNR). The NeRF-based Instant-NGP (28.12 PSNR) lags significantly behind the Gaussian Splatting methods. These scores, which are lower than those for the synthetic Diffuse category, confirm that our real-world captures, with their complex materials and lighting, pose a significant challenge.

\begin{table}[htbp]
  \centering
  \caption{Surface Reconstruction performance on the real-world dataset. Metric: Chamfer Distance ($\downarrow$). Lower is better.}
  \begin{tabular}{@{}lc@{}}
    \toprule
    \textbf{Method} & \textbf{Real-World Dataset} \\
    \midrule
    Instant-NGP~\cite{muller2022instant}     & 0.139 \\
    Neus2~\cite{wang2023neus2}      & 0.132 \\
    2DGS~\cite{huang20242d}            & 0.105 \\
    PGSR~\cite{chen2025pgsr}            & 0.207 \\
    \bottomrule
  \end{tabular}
  \label{tab:surface_reconstruction_performance_real}
\end{table}

\noindent \textbf{Surface Reconstruction Analysis.} The surface reconstruction results in Table~\ref{tab:surface_reconstruction_performance_real} highlight the significant challenge our real-world dataset poses for all tested methods. The results demonstrates the similar  phenomenon to those on the synthetic dataset, indicating that the  SOTA methods, designed for simple Lambertian surfaces, are not robust to the complex, non-Lambertian challenges our dataset exposes.

\section{Relighting}
\label{sup-appendix:relighting}

We evaluated four SOTA relighting methods on the \method dataset. These methods are designed to decompose materials into their intrinsic properties (albedo, roughness, etc.) to allow for rendering under novel lighting conditions. We report the average PSNR and SSIM for novel-light rendering against the ground-truth images.

\begin{table}[htbp]
  \centering
  \caption{Quantitative comparison of SOTA relighting baselines on the \method dataset. }
  \resizebox{\columnwidth}{!}{
  \begin{tabular}{lcccc}
    \hline
    Metrics & GS-IR~\cite{liang2024gs} & GI-GS~\cite{chen2024gi} & NVDiffrec~\cite{munkberg2022extracting} & TensoIR~\cite{jin2023tensoir} \\
    \hline
    PSNR    & 19.32 & 20.02 & 16.98     & 23.39   \\
    SSIM    & 0.812 & 0.826 & 0.795     & 0.874   \\
    \hline
  \end{tabular}
  }
  \label{tab:relighting_results}
\end{table}

The results in Table~\ref{tab:relighting_results} show that TensoIR achieves the best performance among the baselines, with a PSNR of 23.39. The other Gaussian-based methods, GI-GS and GS-IR, perform moderately, while NVDiffrec struggles significantly. However, the modest scores of all methods, with the top performer failing to exceed 24 dB PSNR, confirm the findings from our main paper: the complex, non-Lambertian materials in \method pose a significant challenge for current SOTA relighting techniques. We thus believe our large-scale dataset of physically-based assets will be a valuable resource for driving future research in this area.


\section{Detailed Analysis of Material Parameter Impact}
\label{sup-appendix:material_parameter_impact}

To provide a comprehensive visual reference for the parameter sweep analyzed in the main text, Figure~\ref{fig:material_parameters_collections} presents the rendered appearance of all 48 unique material configurations tested in our experiments. The 48 combinations are structured into three distinct physical categories. Each category contains 16 variations derived from a 4x4 grid spanning Roughness (0.0, 0.3, 0.6, 0.9) and IOR (1.0, 1.3, 1.6, 1.9):

\begin{itemize}
    \item Opaque Non-Metals (Dielectrics): (Metallic=0, Transmission=0). 16 combinations.
    \item Opaque Metals (Conductors): (Metallic=1, Transmission=0). 16 combinations.
    \item Transparent Non-Metals (Dielectrics): (Metallic=0, Transmission=1). 16 combinations.
\end{itemize}
The physically implausible combination of (Metallic=1, Transmission=1) was excluded, resulting in the 48 total test cases. This figure allows for a direct visual correlation between a material's appearance and its corresponding reconstruction quality shown in the main paper's analysis.

\begin{figure}[h!]
    \centering
    \includegraphics[width=\linewidth]{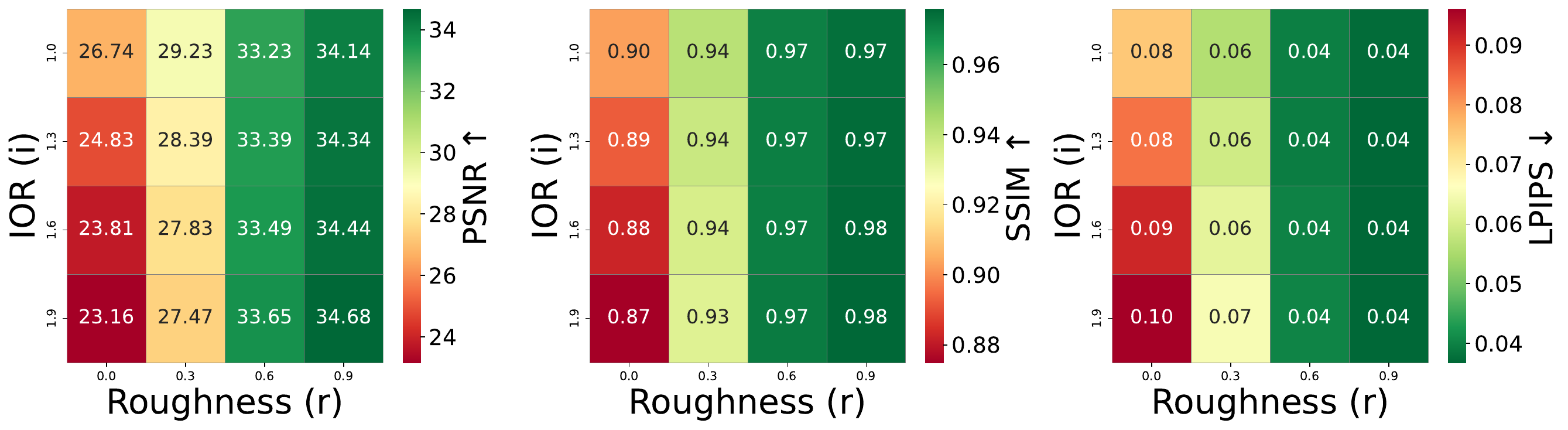}
    \caption{Detailed Impact of Roughness and IOR on Reconstruction Quality for Opaque, Non-Metallic Materials.}
    \label{fig:material_heatmap}
\end{figure}

The heatmaps in Figure~\ref{fig:material_heatmap} provide a granular analysis of the 16 parameter combinations for opaque, non-metallic materials (Metallic=0, Transmission=0). The three heatmaps plot reconstruction quality metrics against Roughness (x-axis) and Index of Refraction (y-axis). This analysis reveals two key insights:

\noindent \textbf{Roughness is the Dominant Factor for Opaque, Non-Metallic Materials.} A strong, consistent trend is visible across all three metrics: reconstruction quality fails catastrophically at low roughness and improves dramatically as roughness increases. At Roughness=0.0, PSNR values are poor, clustering in the 23-27 dB range. As roughness increases to 0.9, the PSNR values improve significantly to the 34-35 dB range.
This confirms the hypothesis from the main text: smooth, low-texture surfaces starve the 3DGS algorithm of the high-frequency features needed for multi-view correspondence. As roughness increases, the material's microsurface scatters light more diffusely, which effectively acts as a high-frequency texture that the algorithm can successfully use for matching, drastically reducing failure.

\noindent \textbf{IOR has a Minimal Effect in Opaque, Non-Metallic Cases.}
In contrast to the strong influence of roughness, the IOR has a very weak, almost negligible, impact on reconstruction quality. For a non-metallic, opaque material, the IOR's primary physical effect is controlling the intensity of specular reflections (the Fresnel effect). Across all metrics, the vertical columns in the heatmaps are nearly uniform in color. For example, at a Roughness of 0.3, the PSNR only varies from 27.47 to 29.23 (a $<2$ dB difference) as IOR spans its entire 1.0-1.9 range. At high Roughness=0.9, the IOR has virtually no impact, with PSNR remaining static between 34.14 and 34.68. This is because the high roughness diffuses all reflections, rendering the IOR-driven Fresnel effect imperceptible.

This detailed breakdown confirms that the reconstruction failures in this subset are driven almost entirely by the lack of geometric features on smooth surfaces, not by the view-dependent reflectivity introduced by IOR. This stands in stark contrast to the transparent and metallic cases, where reflectivity and refraction are the primary causes of failure.

\begin{figure}[h!]
 \centering
 \includegraphics[width=\linewidth]{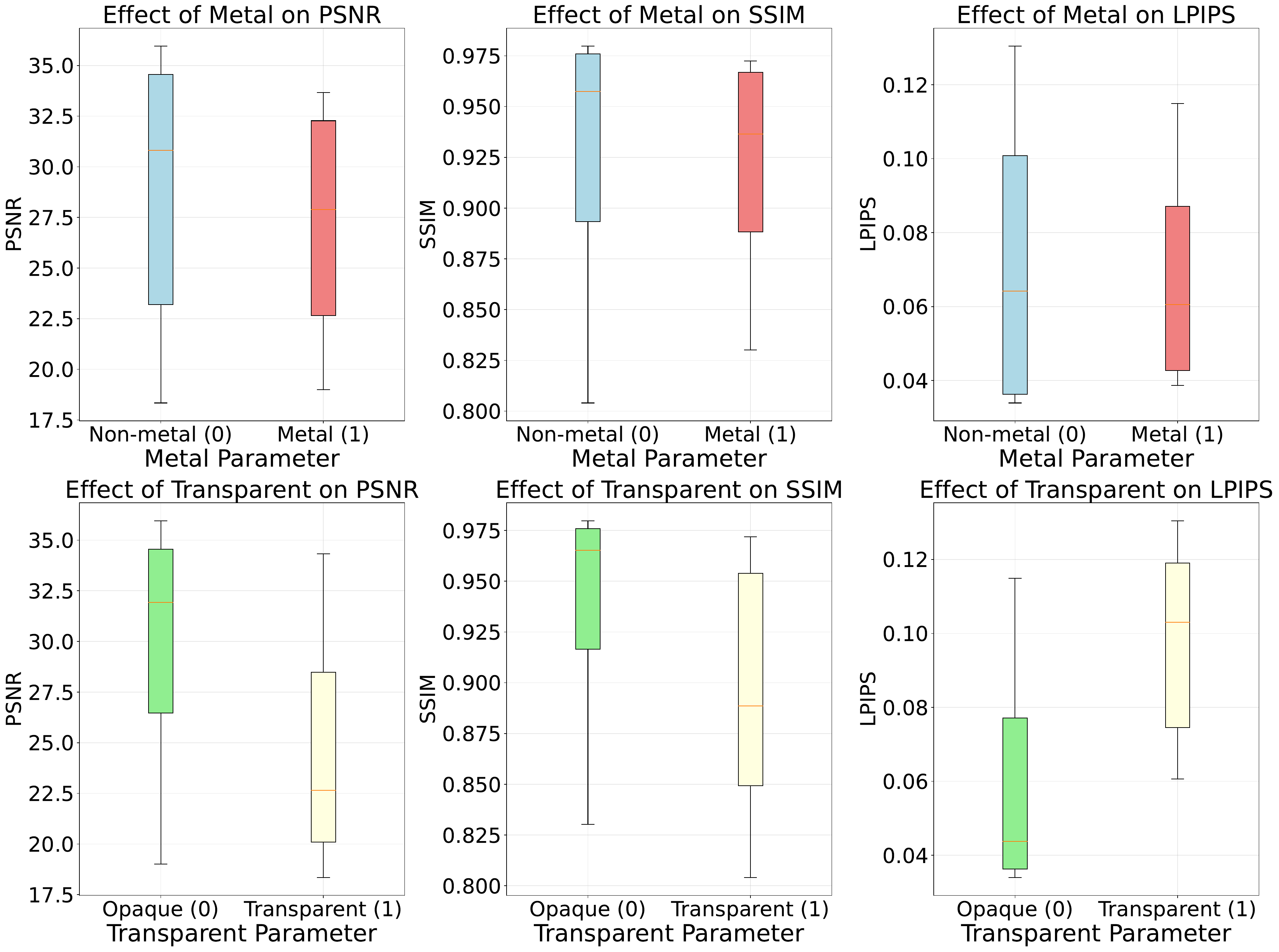}
 \caption{Comparative analysis of reconstruction quality for metallic vs. non-metallic and transparent vs. opaque materials. The box plots show the distribution of PSNR, SSIM, and LPIPS results when aggregating all other material variations.}
\label{fig:material_plots}
\end{figure}

\noindent \textbf{Analysis of Binary Parameter Effects (Metal \& Transparent).}
While the heatmaps analyze the non-metallic, opaque subset, Figure~\ref{fig:material_plots} analyzes the aggregated impact of the binary `Metal' and `Transparent' parameters across all 48 configurations.

\noindent \textbf{Metallic materials cause significant pixel-wise failure.} Setting `Metal=1' has a severe negative impact on reconstruction quality, causing a dramatic drop in median PSNR (from ~33 dB to ~25 dB) and SSIM (from ~0.96 to ~0.91).Interestingly, this degradation is not reflected in the perceptual LPIPS metric, where the median error for metallic objects (~0.062) is slightly better than for non-metallic ones (~0.064).

\noindent \textbf{Transparent materials consistently degrade all metrics.} Transparency (`Transparent=1') is a more consistent failure case, degrading performance across all metrics. It causes a clear drop in median PSNR (from ~30 dB to ~28 dB) and SSIM (from ~0.95 to ~0.92). This negative effect is most pronounced in the perceptual LPIPS metric, where the median error for transparent objects (~0.08) is significantly worse than for opaque objects (~0.05).

\section{A Physically-Based Analysis of Failure Modes in Multi-View 3D Reconstruction}
\label[appendix]{sup-appendix:root cause analysis}

\subsection{The Physics of Image Formation: A Ground-Truth Model}
To comprehend why certain algorithms fail, one must first establish a ground-truth model of the physical process they attempt to approximate. In computer graphics and physics, the interaction of light with surfaces is meticulously described by the principles of radiometry and physically based rendering. This section establishes a comprehensive and mathematically rigorous model of image formation, which will serve as the physical reality against which the simplified models used in computer vision are compared and critiqued.

\textbf{Formulate Light in Equilibrium.}
The cornerstone of physically based rendering is the Rendering Equation, an integral equation independently introduced by Immel et al.,\cite{A_radiosity_method_for_non-diffuse_environments} and Kajiya et al.,\cite{kajiya1986rendering} in 1986. It provides a complete and elegant description of the equilibrium state of light transport in a scene, defining the amount of light leaving any given point on a surface in any given direction. The equation is a statement of energy conservation, asserting that the total light leaving a point is the sum of the light it emits and the light it reflects from all other sources in the environment.

Its canonical form is expressed as:
\begin{equation}
L_o(x, \omega_o) = L_e(x, \omega_o) + \int_{\Omega} f_r(x, \omega_i, \omega_o) L_i(x, \omega_i) (\mathbf{n} \cdot \omega_i) d\omega_i
\end{equation}

Each component of this equation has a precise physical meaning:
\begin{itemize}
    \item $L_o(x, \omega_o)$: The outgoing radiance from a point $x$ on a surface in a specific direction $\omega_o$. Radiance is the radiometric quantity of light energy per unit solid angle per unit projected area, and it is what a camera sensor ultimately measures to form an image.
    \item $L_e(x, \omega_o)$: The emitted radiance from point $x$ in direction $\omega_o$. This term is non-zero only for surfaces that are light sources themselves. For most objects in a scene, this term is zero.
    \item $\int_{\Omega}$: An integral over the unit hemisphere $\Omega$ oriented around the surface normal $\mathbf{n}$ at point $x$. This signifies that to calculate the total reflected light, one must account for all possible incoming light directions from the entire hemisphere above the surface.
    \item $f_r(x, \omega_i, \omega_o)$: The Bidirectional Reflectance Distribution Function (BRDF). This function is the heart of material appearance, defining the ratio of reflected radiance in the outgoing direction $\omega_o$ to the incident irradiance from an incoming direction $\omega_i$.It mathematically describes the intrinsic reflective properties of the material at point $x$.
    \item $L_i(x, \omega_i)$: The incident radiance arriving at point $x$ from direction $\omega_i$. This term is what makes the Rendering Equation a global and recursive construct. The light arriving at point $x$ is simply the outgoing light, $L_o$, from some other point in the scene that is visible from $x$ along the direction $-\omega_i$. This recursive definition means that the appearance of a single point is dependent on the appearance of every other point in the scene, modeling phenomena like indirect illumination and color bleeding.
    \item $(\mathbf{n} \cdot \omega_i)$: Lambert's Cosine Law. This is a geometric term representing the dot product between the surface normal $\mathbf{n}$ and the incoming light direction $\omega_i$. It accounts for the fact that a surface receives less light flux per unit area from sources at grazing angles, as the incident energy is spread over a larger area.
\end{itemize}

The recursive and integral nature of the Rendering Equation reveals a fundamental truth about image formation: it is a global phenomenon. The color of a single pixel is not a purely local property but is the result of a complex interplay of light bouncing throughout the entire scene, converging at that point before traveling to the camera. This global light transport system, which includes inter-reflections between surfaces, is a physical reality that most local, patch-based computer vision algorithms fundamentally fail to model.

\subsection{Light Behavior with Complex Materials}
\label[appendix]{sup-appendix:light_behavior}
A material’s electronic response, internal composition, and surface microgeometry dictate the fate of impinging light—whether it is reflected, refracted, transmitted, or absorbed. In rendering and inverse‐vision, these processes are commonly expressed via bidirectional scattering functions and Fresnel’s laws~\cite{judd1942fresnel, lvovsky2013fresnel}.

\subsubsection{Fresnel Reflectance and Refraction}  
At the interface between media with refractive indices $n_1$ and $n_2$, the proportions of reflected and refracted light are governed by Fresnel’s equations~\cite{judd1942fresnel}. For unpolarized light, the reflectance $F_r$ as a function of incident angle $\theta_i$ is
\begin{equation}
F_r(\theta_i) 
= \tfrac{1}{2}\Bigl[\bigl(\tfrac{n_1\cos\theta_i - n_2\cos\theta_t}{n_1\cos\theta_i + n_2\cos\theta_t}\bigr)^{2}
+ \bigl(\tfrac{n_2\cos\theta_i - n_1\cos\theta_t}{n_2\cos\theta_i + n_1\cos\theta_t}\bigr)^{2}\Bigr]
\end{equation}
with $\theta_t$ given by Snell’s law $n_1\sin\theta_i = n_2\sin\theta_t$~\cite{Greivenkamp2004}. The transmitted fraction $T$ then satisfies energy conservation $F_r + T + A=1$, where $A$ is absorption.

\subsubsection{Microfacet BRDF for Specular Reflection}  
Metals and glossy dielectrics exhibit specular reflection that varies with surface roughness. The microfacet BRDF~\cite{CookTorrance1982} is
\begin{equation}
f_r(\omega_i,\omega_o)
= \frac{D(h)\,G(\omega_i,\omega_o)\,F_r(\omega_i\cdot h)}{4\cos\theta_i\,\cos\theta_o}\,
\end{equation}
where $\omega_i,\omega_o$ are the incident and exitant directions, $h$ is the half‐vector, $D$ the normal distribution function, $G$ the geometric shadowing–masking term, and $F_r$ the Fresnel term. We adopt the Trowbridge–Reitz (GGX) ~\cite{trowbridge1975average} distribution
\begin{equation}
D(h)=\frac{\alpha^2}{\pi\bigl[(\alpha^2-1)\cos^2\theta_h +1\bigr]^2}
\end{equation}
and the Smith–Walter $G$ function~\cite{Heitz2014,Walter2007}.  Figure~\ref{fig:specular_reflection_explain} illustrates the process.

\begin{figure}[h!]
    \centering
    \includegraphics[width=\linewidth]{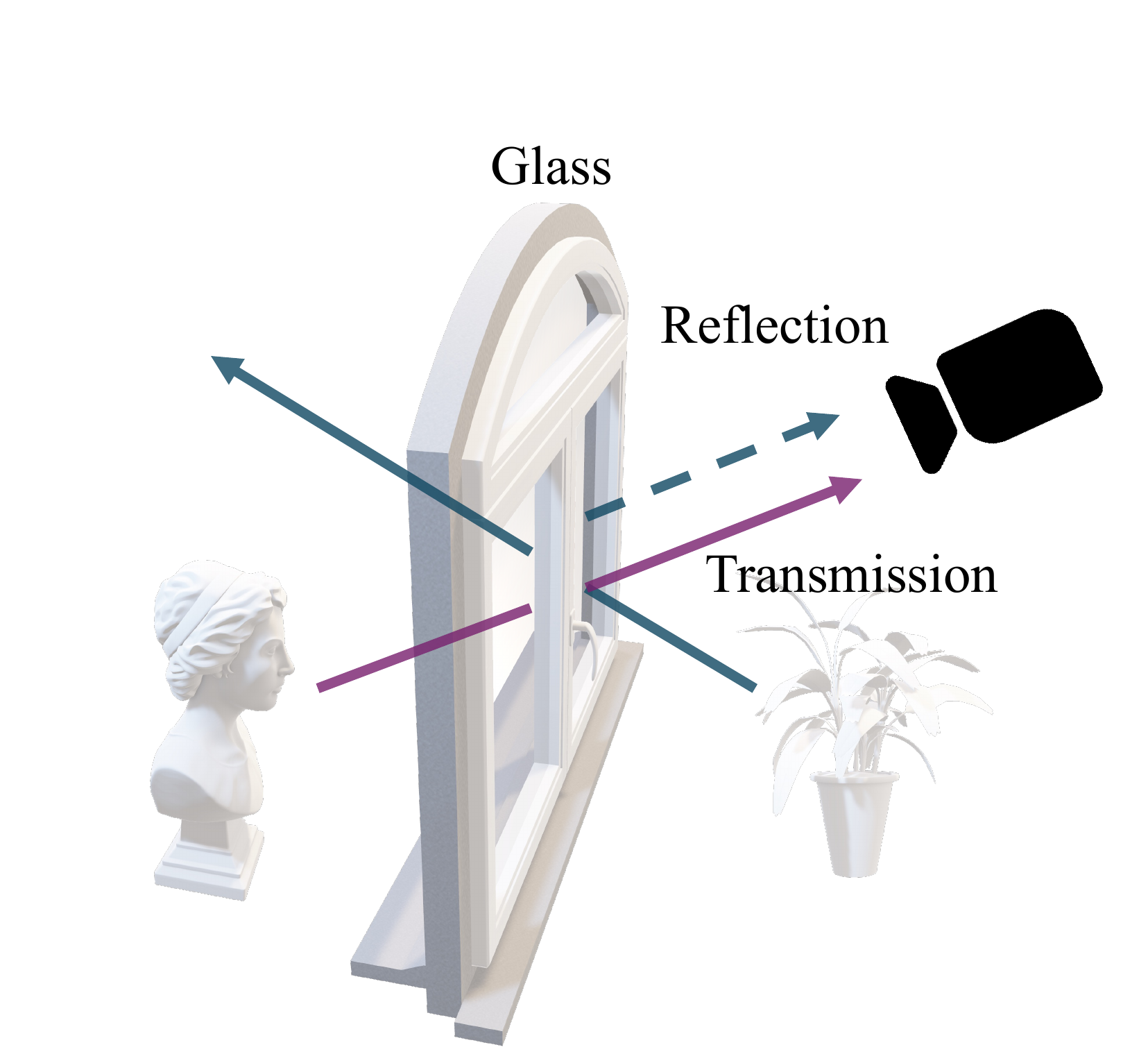}
    \caption{Specular Reflection}
    \label{fig:specular_reflection_explain}
\end{figure}

\subsection{BTDF for Transmission}  
Transparent dielectrics require a bidirectional transmission distribution function (BTDF) to model refracted light. A common form combines Fresnel‐weighted refraction with microfacet masking~\cite{Walter2007}:
\begin{equation}
f_t(\omega_i,\omega_o)
= \frac{(1 - F_r)\,D(h)\,G(\omega_i,\omega_o)\,(n_2/n_1)^2}
{4\cos\theta_i\,\cos\theta_o}\
\end{equation}

\subsection{Diffuse Scattering and Absorption}  
Pigmented or rough materials scatter light diffusely. Lambert’s law approximates uniform scattering:
\begin{equation}
f_d = \frac{\rho}{\pi}
\end{equation}
where $\rho$ is albedo. Subsurface scattering in plastics and paints can be modeled via the Kubelka–Munk theory~\cite{KubelkaMunk1931} or dipole diffusion~\cite{Jensen2001}.

\begin{table*}[ht!]
\centering
\caption{Mapping of Material Properties to Violated Algorithmic Assumptions.}
\begin{tabular}{l|p{4cm}|p{4cm}|p{4cm}}
\toprule
\textbf{Material Type} & \textbf{Primary Phenomenon} & \textbf{Violated Assumption(s)} & \textbf{Consequence} \\
\midrule
Low-Texture & Lack of high-frequency albedo variation. & Feature Correspondence, Photometric Uniqueness & SfM fails to find matches. MVS cost volume is ambiguous, leading to noisy/flat geometry. \\ 

\hline

Reflective & View-dependent specular reflection (BRDF). & Photometric Consistency & SfM matching can be unreliable. MVS produces noisy, inaccurate, or incomplete geometry. \\ 

\hline

Transparent & Refraction of light (Snell's Law), internal reflections. & Photometric Consistency, Linear Light Propagation (Epipolar Geometry) & Complete failure. Photometric matching is impossible. Triangulation is geometrically invalid. \\
\bottomrule
\end{tabular}
\label{tab:failure_modes}
\end{table*}

\subsection{Foundational Assumptions in Multi-View Reconstruction}
Standard reconstruction pipelines, composed of Structure-from-Motion (SfM) and Multi-View Stereo (MVS), are built upon core assumptions that simplify the complex physics of light into a tractable problem.

\subsubsection{The Assumption of Photometric Consistency in MVS}
The central assumption of MVS is that a true 3D point on a surface will exhibit a similar color across multiple camera views. MVS algorithms construct a cost volume by comparing image patches between views at hypothesized depths, seeking the depth that minimizes the matching cost (e.g., Sum of Squared Differences). This assumption of view-invariant appearance is, in essence, an implicit assumption of Lambertian reflectance. Any deviation from this ideal diffuse behavior represents a potential violation of this core assumption.

\subsubsection{The Assumption of Feature Correspondence in SfM}
SfM pipelines operate by detecting and matching sparse, salient feature points (e.g., using SIFT~\cite{lowe2004distinctive}) across multiple views to solve for camera poses. This relies on the assumption that the local appearance of a feature is sufficiently stable across viewpoints to allow for reliable matching. This assumption breaks down under the drastic, non-linear appearance changes caused by strong specular reflections.

\subsubsection{The Assumption of Linear Light Propagation in Epipolar Geometry}
The entire geometric framework of multi-view reconstruction is predicated on a simple and fundamental assumption: light travels in a straight line from a 3D point to the camera center.Any phenomenon that causes the light path to bend, such as refraction, will invalidate the principles of epipolar geometry and the triangulation methods used to compute 3D structure.

\subsection{Conclusion and Future Directions.}

The failures of conventional multi-view 3D reconstruction pipelines on reflective, low-texture, and transparent materials are not isolated algorithmic bugs. They are the direct and predictable consequences of a fundamental conflict between the simplified physical models embedded in these algorithms and the complex reality of light transport. The entire SfM-MVS pipeline is built on a set of assumptions that hold only for a small subset of real-world scenes: those that are well-textured, opaque, and largely diffuse. The core argument can be summarized by mapping material properties to the specific assumptions they violate in Table~\ref{tab:failure_modes}.


\section{Annotations for Generative 3D Vision Tasks}
\label[appendix]{sup-appendix:generation_annotations}

We extend \method beyond perception tasks by providing detailed textual annotations for each instance, enabling future research in generative 3D vision. This section describes our annotation methodology, format, and provides comprehensive examples to facilitate integration with downstream generation pipelines.

\subsection{Annotation Methodology}

Our annotation pipeline leverages the Qwen3-VL-30B-A3B-Instruct~\cite{qwen3technicalreport} vision model to generate structured descriptions for each instance in \method. The annotation process captures four key aspects:

\begin{enumerate}
    \item Detailed Material Descriptions: Comprehensive descriptions of surface properties, including reflectance characteristics, roughness, transparency, and material composition.
    \item Lighting Condition Tags: Explicit annotations of lighting setup, including light types, directions, intensities, and environmental illumination.
    \item Semantic Instance Descriptions: Object-level descriptions that capture both geometric and appearance properties relevant for 3D generation.
\end{enumerate}

\subsection{Annotation Format}

Each instance in \method is annotated with structured text following a hierarchical schema. The format includes separate fields for material properties, lighting descriptions, and a natural language generation description which can be used as prompts for downstream tasks.

\begin{figure}[t]
\centering
\includegraphics[width=0.9\linewidth]{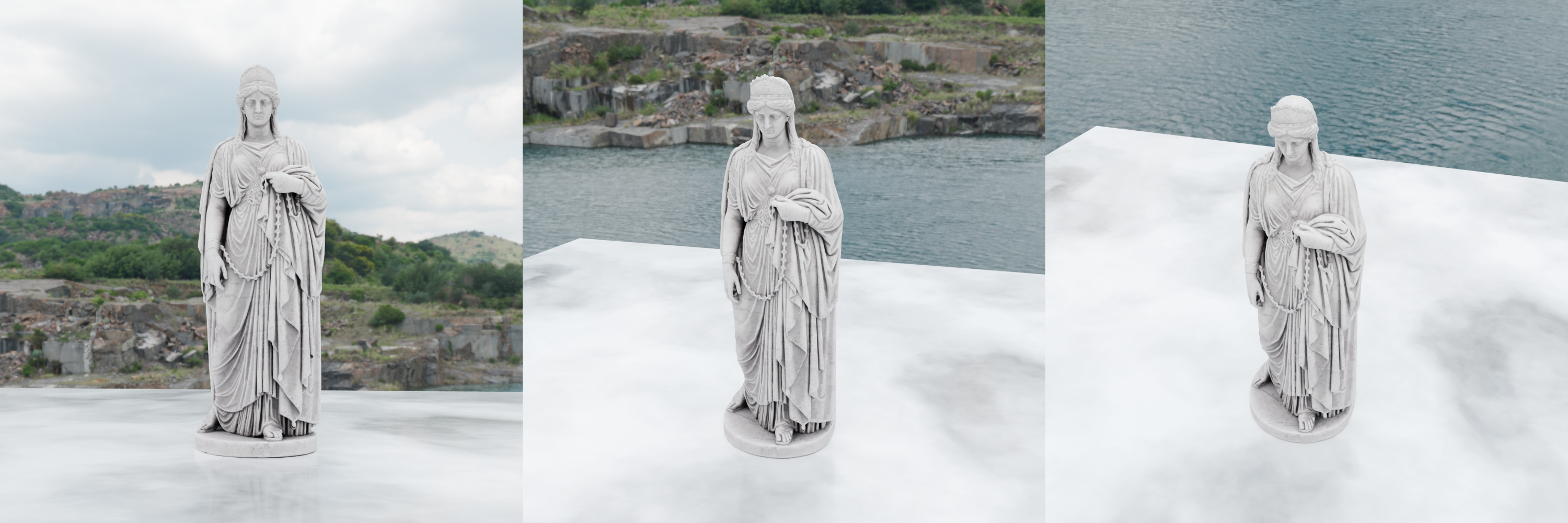}
\caption{Input images used for annotation (low/middle/high camera angles).}
\label{fig:input_images}
\end{figure}

\begin{figure*}[t]
\centering
\begin{tcolorbox}[
  title=Asset Image Annotation Prompt,
  width=\linewidth,
  colback=white, colframe=black, colbacktitle=black,
  enhanced, center,
  attach boxed title to top left={yshift=-0.1in,xshift=0.15in},
  boxed title style={boxrule=0pt,colframe=white}
]

\textbf{Prompt:}

You are a rigorous visual annotator. Please output a concise, structured result \textbf{containing only JSON} based on three images of the same instance (low/flat/high angles).

\begin{itemize}
    \item The model is placed on a marble surface.
    \item The camera height ratio for the three angles are 0.3 / 1.0 / 1.5 (relative to the model height).
    \item category.main/sub must be \textbf{strictly copied} from meta.json; no reclassification is allowed.
    \item Only judge "material appearance attributes" and "environment"; do not output database fields (hasGlass, transparent, isGenerated, material\_name).
    \item All values must be selected from the provided enums or value ranges.
\end{itemize}

Note: Output Fields (Only These!)
\begin{itemize}
    \item instance\_id: string (from meta.json or parameters)
    \item category: \{ main: string, sub: string \} (strictly copied from meta.json; reclassification is prohibited)
    \item material\_properties:
    \begin{itemize} 
        \item glossiness: ``matte"|``semi-gloss"|``glossy"|``mirror"
        \item roughness: number (0.0-1.0)
        \item reflectivity: ``none"|``weak"|``medium"|``strong"
        \item texture\_scale: "fine"|``medium"|``coarse"
        \item anisotropy: ``none"|``weak"|``strong"
        \item metallic\_hint: ``non-metal"|``mixed"|``metallic"
    \end{itemize}
    \item environment:
    \begin{itemize}
        \item indoor\_outdoor: ``indoor"|``outdoor"|``unknown"
        \item light\_type: ``natural"|``artificial"|``mixed"|``unknown"
        \item light\_intensity: ``weak"|``medium"|``strong"|``unknown"
        \item key\_light\_direction: ``left"|``right"|``front"|``back"|``top"|``ambient"|``unknown"
        \item shadow\_hardness: ``soft"|``medium"|``hard"|``unknown"
    \end{itemize}
    \item description: string (briefly describe the appearance of the instance in English)
\end{itemize}

Note: Category Fields to Be Strictly Copied!
\begin{itemize}
    \item category.main = ``\{cat\_main\}"
    \item category.sub  = ``\{cat\_sub\}"
\end{itemize}

The following is meta.json (for reference only; do not use its database fields for judgment):

\begin{verbatim}
{
    "instance_id": "7db354ab-34a0-4634-b4e3-cf100a4736c9",
    "main_category": "culture",
    "sub_category": "sculpture",
    "model_name": "Roman_Statue.blend_Zenobia in Chains statue",
    "material_name": "cleanSnow",
    "env_name": "quarry_cloudy_4k.exr",
    "hasGlass": false,
    "isGenerated": false,
    "transparent": false
}
\end{verbatim}

\end{tcolorbox}
\caption{The prompt for generating description for an asset}
\label{fig:prompt}
\end{figure*}

\begin{figure*}[t]
\begin{tcolorbox}[
  title=Generated Annotation Output (tags.json),
  width=\linewidth,
  colback=white, colframe=black, colbacktitle=black,
  enhanced, center,
  attach boxed title to top left={yshift=-0.1in,xshift=0.15in},
  boxed title style={boxrule=0pt,colframe=white}
]

This is the resulting \textit{tags.json} file generated by the annotation pipeline, corresponding to the input from Figure~\ref{fig:prompt}.

\begin{verbatim}
{
  "instance_id": "7db354ab-34a0-4634-b4e3-cf100a4736c9",
  "category": {
    "main": "culture",
    "sub": "sculpture"
  },
  "material_properties": {
    "glossiness": "matte",
    "roughness": 0.3,
    "reflectivity": "weak",
    "texture_scale": "fine",
    "anisotropy": "none",
    "metallic_hint": "non-metal"
  },
  "environment": {
    "indoor_outdoor": "outdoor",
    "light_type": "natural",
    "light_intensity": "medium",
    "key_light_direction": "top",
    "shadow_hardness": "soft"
  },
  "description": "A classical marble sculpture of a robed female figure, 
  standing on a white marble platform. The statue has a matte, light gray 
  finish with fine surface details. It is positioned outdoors in a quarry-like
  environment with a body of water and rocky hills in the background. 
  The lighting is soft and diffused, suggesting an overcast sky."
}
\end{verbatim}

\end{tcolorbox}
\caption{The structured \textit{tags.json} output from our annotation pipeline. This includes the strictly copied category, the VLLM-inferred material and environment properties, and the natural language description.}
\label{fig:annotation_output}
\end{figure*}

\subsection{Annotation Examples}

We provide a concrete example of our annotation pipeline in Figure~\ref{fig:input_images}, Figure~\ref{fig:prompt} and Figure~\ref{fig:annotation_output}.

Figure~\ref{fig:input_images} and 
Figure~\ref{fig:prompt} show the images and complete prompt used to instruct the VLLM, along with a sample \textit{meta.json} file. This input file provides database information, such as \textit{material\_name}, and the ground-truth categories.

Figure~\ref{fig:annotation_output} shows the corresponding \textit{tags.json} file generated by our pipeline. As demonstrated, the process strictly adheres to the prompt's rules: it correctly copies the \textit{category} field from the input and populates the \textit{material\_properties}, \textit{environment}, and \textit{description} fields based on the model's visual analysis.

\subsection{Integration with Generation Pipelines}

Our rich annotations enable seamless integration with various generative 3D vision pipelines. We discuss specific use cases for different generation tasks.

\subsubsection{Text-to-3D Generation}

The natural language generation prompts can be directly used as conditioning text for diffusion-based 3D generation models~\cite{poole2022dreamfusion, yu2023text}. The material and lighting tags enable:

\begin{itemize}
    \item Material-aware NeRF optimization: Tags guide material parameter initialization and constraints during neural radiance field training
    \item PBR parameter prediction: Separate channels for roughness, metallic, and normal maps
    \item Multi-view consistent rendering: Lighting descriptions ensure photometric consistency across views
\end{itemize}

\subsubsection{Text-to-Texture Synthesis}

Material property annotations guide texture generation pipelines~\cite{richardson2023texture, chen2023text2tex}:

\begin{itemize}
    \item Roughness and metallic maps: Surface property map synthesis
    \item Normal map inference: Surface detail generation from descriptions
    \item Environment-aware baking: Lighting-consistent texture synthesis
\end{itemize}

\subsubsection{Image-to-3D Reconstruction}

Lighting descriptions enable advanced reconstruction techniques~\cite{ye2025hi3dgen, liu2024one}:

\begin{itemize}
    \item Relighting capability: Material inference for novel lighting conditions
    \item Lighting-invariant reconstruction: Robust 3D shape recovery under varying illumination
\end{itemize}

\subsubsection{Image Editing and Manipulation}

Rich annotations support material-aware image editing~\cite{nam2024contrastive, li2024zone}:

\begin{itemize}
    \item Material-consistent inpainting: Preserving material properties during completion
    \item Lighting-aware object insertion: Matching illumination of inserted objects
    \item Physical plausibility checking: Validating edits against material-lighting interactions
\end{itemize}

\subsection{Annotation Statistics}

Table~\ref{tab:annotation_stats} provides comprehensive statistics of our annotation dataset, demonstrating the scale and diversity of our annotations for generative tasks.

\begin{table}[t]
\centering
\begin{tabular}{@{}lc@{}}
\toprule
\textbf{Metric} & \textbf{Value} \\
\midrule
Total annotated instances & 126768 \\
Average description length & 268.74 words \\
Unique material & 22 \\
Unique lighting conditions & 2700+ \\
\midrule
\multicolumn{2}{@{}l@{}}{\textit{Material Complexity Distribution:}} \\
\quad Diffuse & 18.2\% \\
\quad Transparent & 9.1\% \\
\quad Metallic & 22.7\% \\
\quad Glossy-Textured & 13.6\% \\
\quad Glossy-Low-Texture & 36.4\% \\
\midrule
\multicolumn{2}{@{}l@{}}{\textit{Lighting Condition Distribution:}} \\
\quad Indoor - Furnished & 8.58\% \\
\quad Indoor - Empty & 16.22\% \\
\quad Outdoor - Natural & 28.24\% \\
\quad Outdoor - Urban & 21.22\% \\
\quad Studio & 25.74\% \\
\bottomrule
\end{tabular}
\caption{Statistics of generation-task annotations in \method. The dataset covers diverse material types and lighting conditions suitable for various generative 3D vision tasks.}
\label{tab:annotation_stats}
\end{table}


\section{Related Works}
\label{sup-appendix:related works}

\subsection{Specular Highlight \& Reflection Removal}
Specular Highlight Removal (SHR) and Single Image Reflection Removal (SIRR) aim to separate interfering light from true scene content. SIRR is a severely ill-posed problem, and modern deep learning approaches are often bottlenecked by an "insufficiency of densely-labeled training data". Recent work like RRW~\cite{zhu2024revisiting} confronts this by creating large-scale, aligned real-world datasets. Architecturally, the field has evolved from location-aware models~\cite{Dong_2021_ICCV} to dual-stream networks, such as the interactive transformers in DSIT~\cite{hu2024single}. The related field of SHR also relies on deep learning, highlighting the importance of "leveraging large-scale synthetic data" for generalization~\cite{Fu_2023_ICCV}.

\subsection{Relighting}

Modern 3D reconstruction with Neural Radiance Fields~\cite{mildenhall2021nerf} excels at view synthesis but entangles geometry, materials, and lighting, which hinders relighting. This "baked-in" problem spurred research into disentangling these properties. Early works like NeRFactor~\cite{zhang2021nerfactor}, and PhySG~\cite{zhang2021physg} factorized the implicit field but remained computationally expensive and often limited to low-frequency lighting. To overcome this, two explicit strategies emerged. First, Munkberg et al., ~\cite{munkberg2022extracting} jointly optimized an explicit triangular mesh, materials, and all-frequency lighting using a differentiable rasterizer and DMTet. Second, the paradigm shifted to 3D Gaussian Splatting, GS-IR~\cite{liang2024gs} adapted inverse rendering to this efficient representation to decompose physical properties. The most recent works, such as GI-GS~\cite{chen2024gi}, now address the limitations of initial 3DGS methods by explicitly modeling global illumination, often using screen-space path tracing to separate direct and indirect lighting.


\begin{figure*}[ht!]
\vspace{-5mm}
    \centering
    \includegraphics[width=\linewidth]{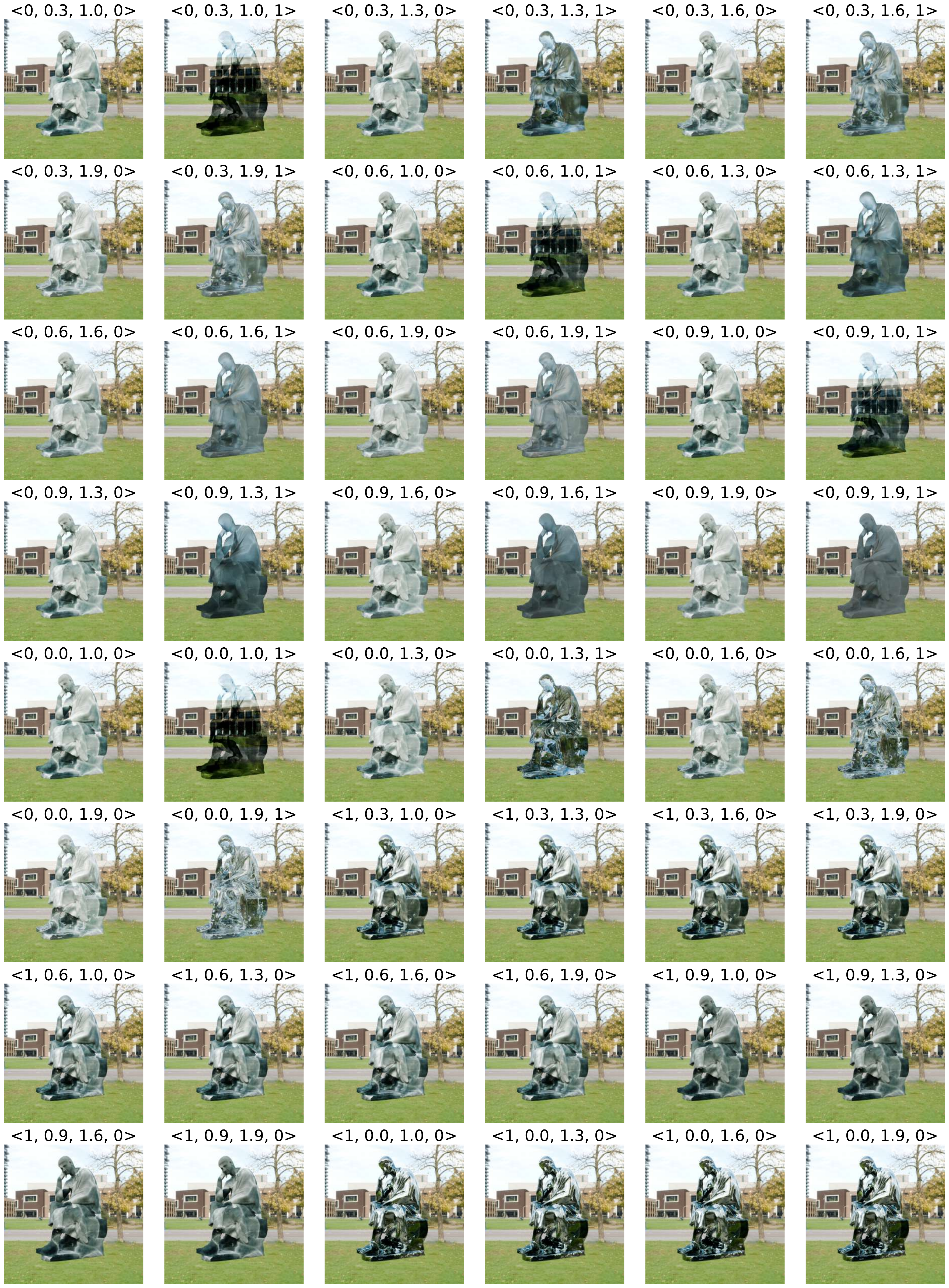}
    \caption{Materials of various parameters show different physical phenomenon. The parameters is in format of $<m, r, i, t>$ }
    \label{fig:material_parameters_collections}
\end{figure*}

\section{More Qualitative Examples}
\label{sup-appendix:more qualitative examples}
We provide additional qualitative examples in this section. Figure~\ref{fig:examples_shapes} showcases various shapes, while Figure~\ref{fig:example_materials} details objects with different materials. For our 3D-generated instances, Figure~\ref{fig:3dgen_shape_material} shows diverse shapes and materials, and Figure~\ref{fig:3dgen_lighting} displays a steel asset under various lighting conditions. Finally, Figure~\ref{fig:more real capture} presents more real-world capture instances.

\begin{figure*}
    \centering
    \includegraphics[width=\linewidth]{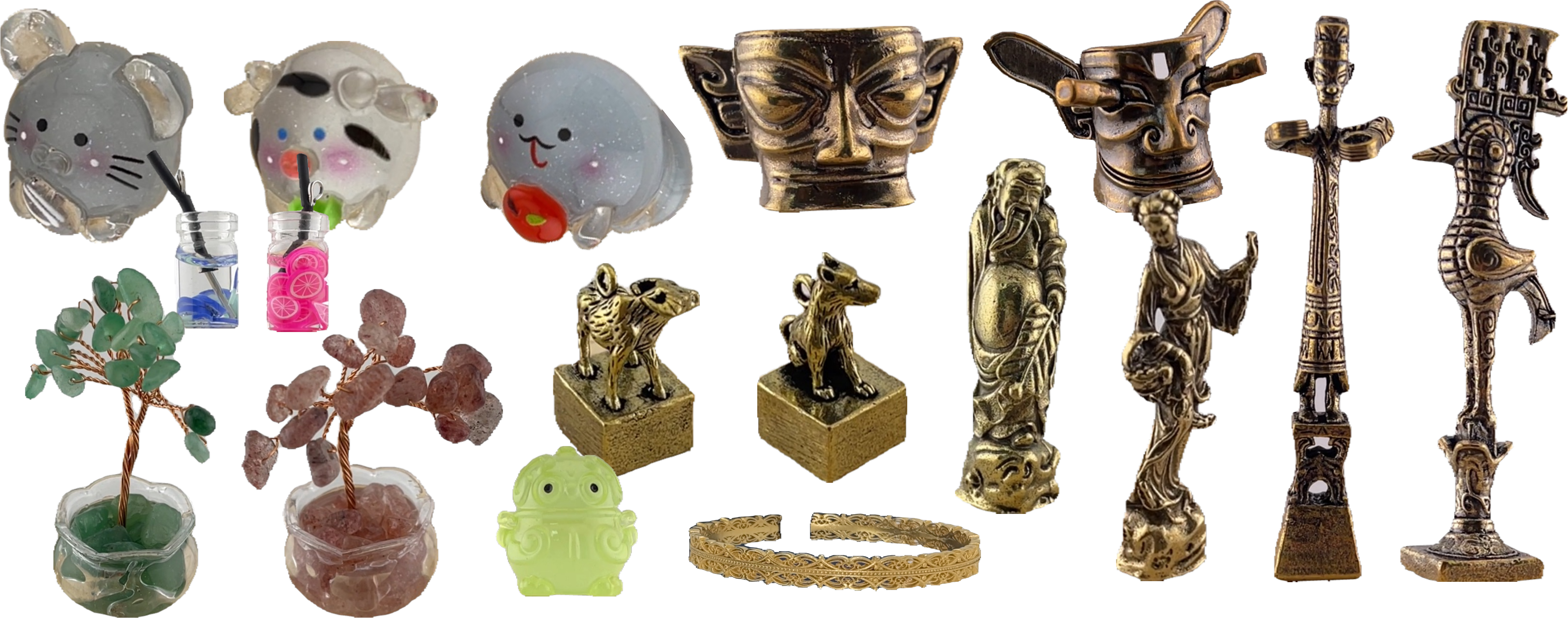}
    \caption{More real-world capture instances, including semi-transparent, reflective, and low-texture objects.}
    \label{fig:more real capture}
\end{figure*}

\begin{figure*}[ht!]
    \centering
    \includegraphics[width=\linewidth]{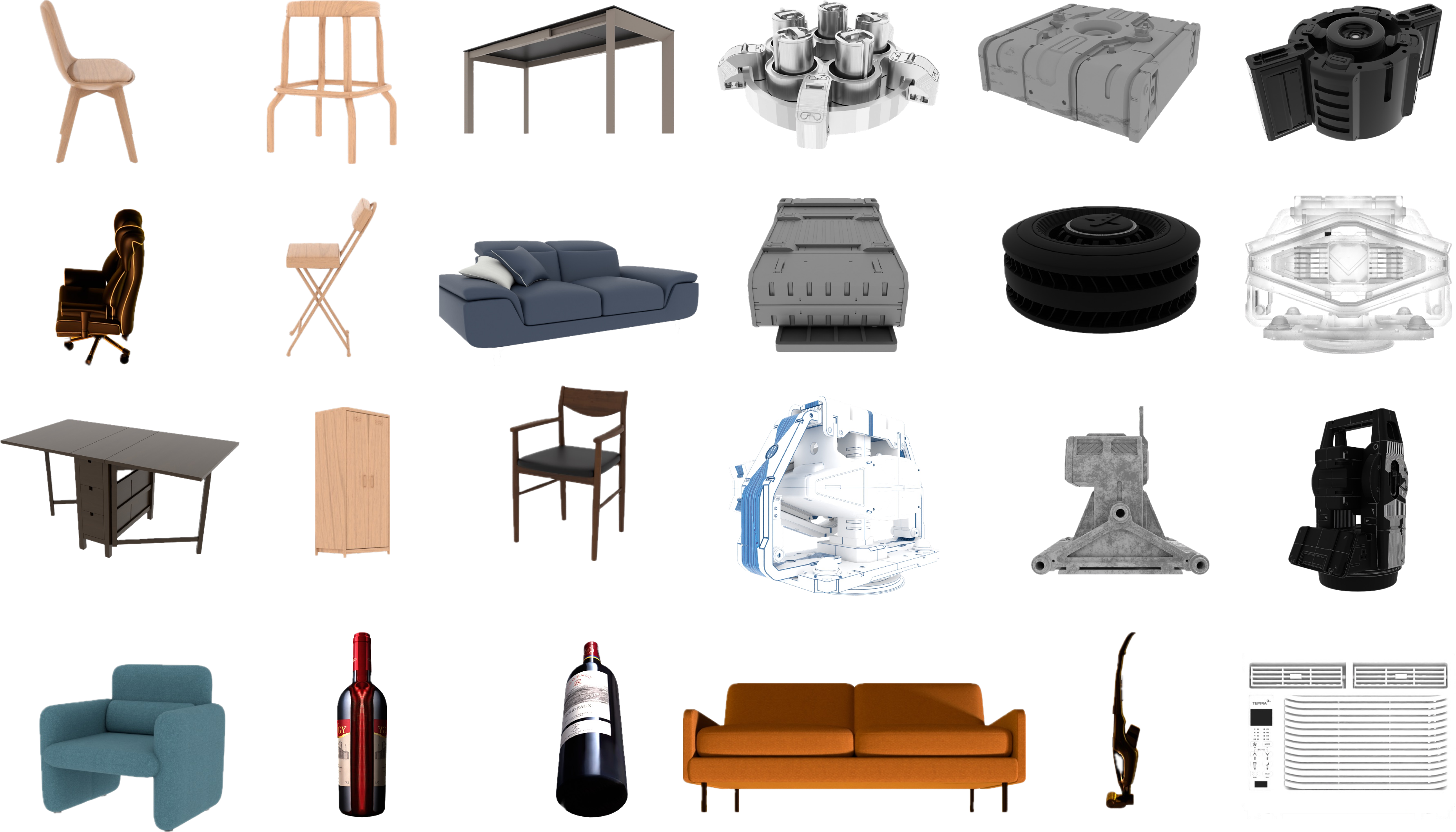}
    \caption{The synthetic objects of various shapes}
    \label{fig:examples_shapes}
\end{figure*}

\begin{figure*}[ht!]
    \centering
    \includegraphics[width=\linewidth]{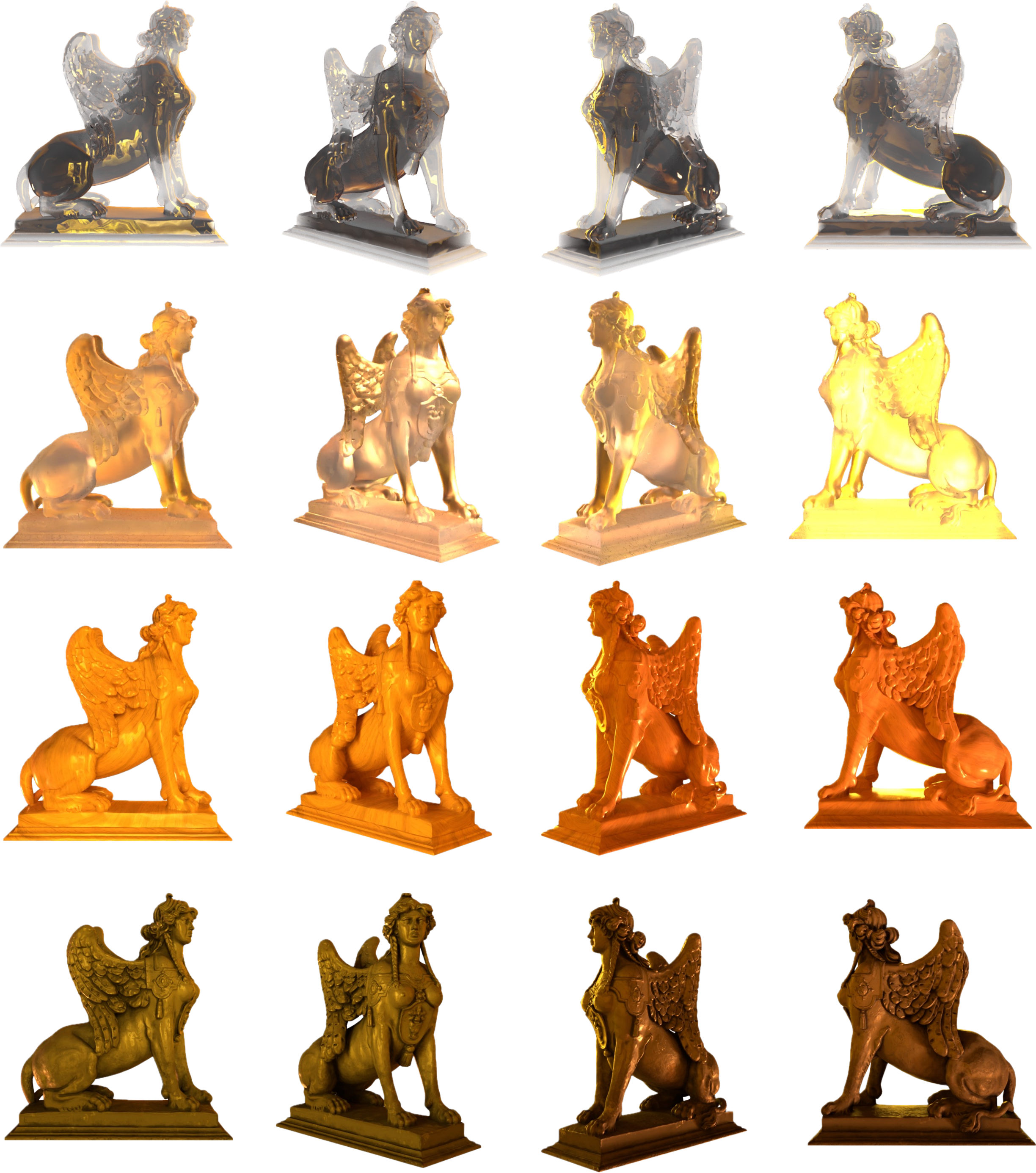}
    \caption{The object with the same shape but made of different materials under identical lighting condition}
    \label{fig:example_materials}
\end{figure*}

\begin{figure*}[ht]
\vspace{-8mm}
    \centering
    \includegraphics[width=\linewidth]{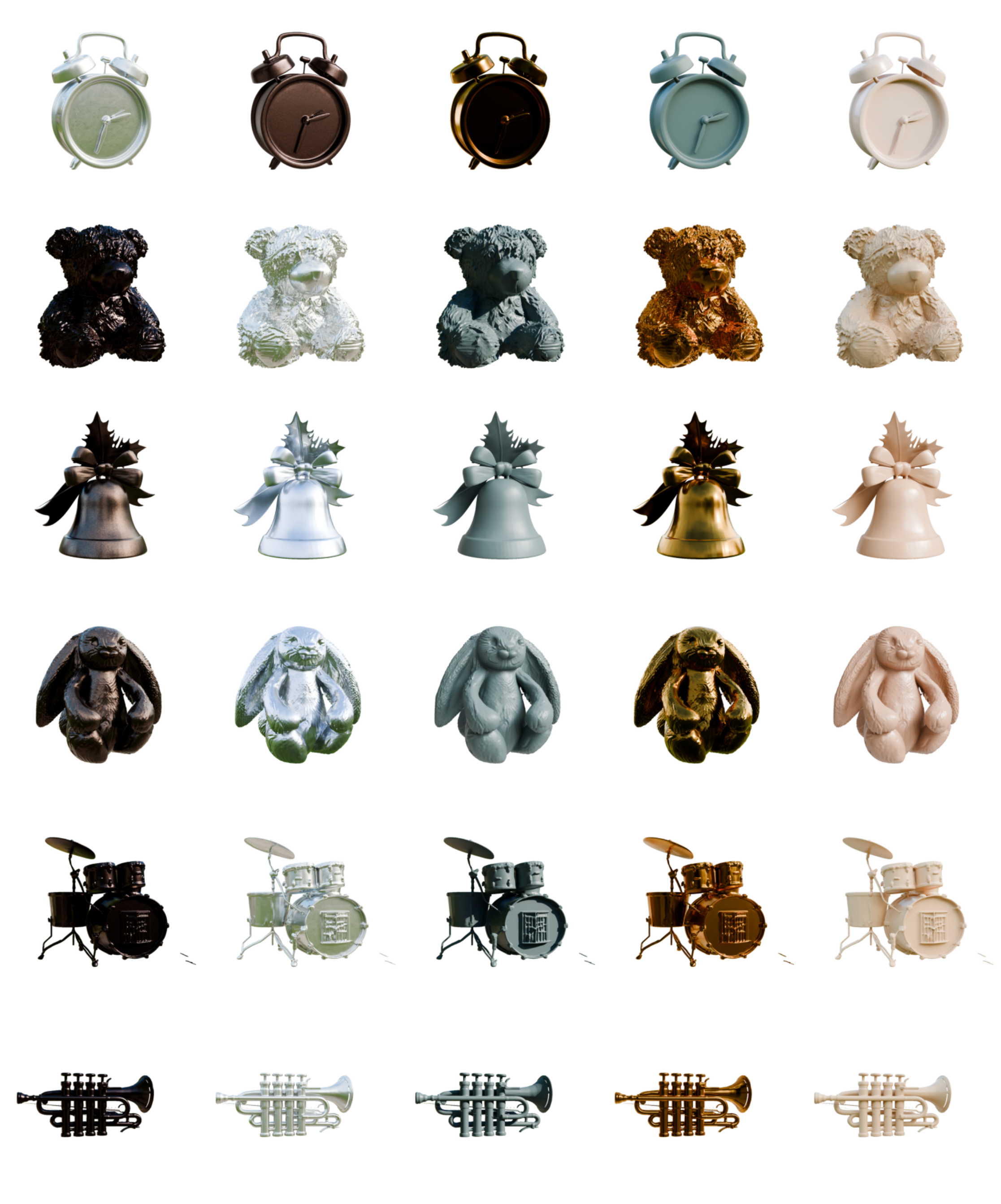}
    \caption{Various shapes of generated 3D assets made of different materials}
    \label{fig:3dgen_shape_material}
\end{figure*}

\begin{figure*}[ht!]
    \centering
    \includegraphics[width=\linewidth]{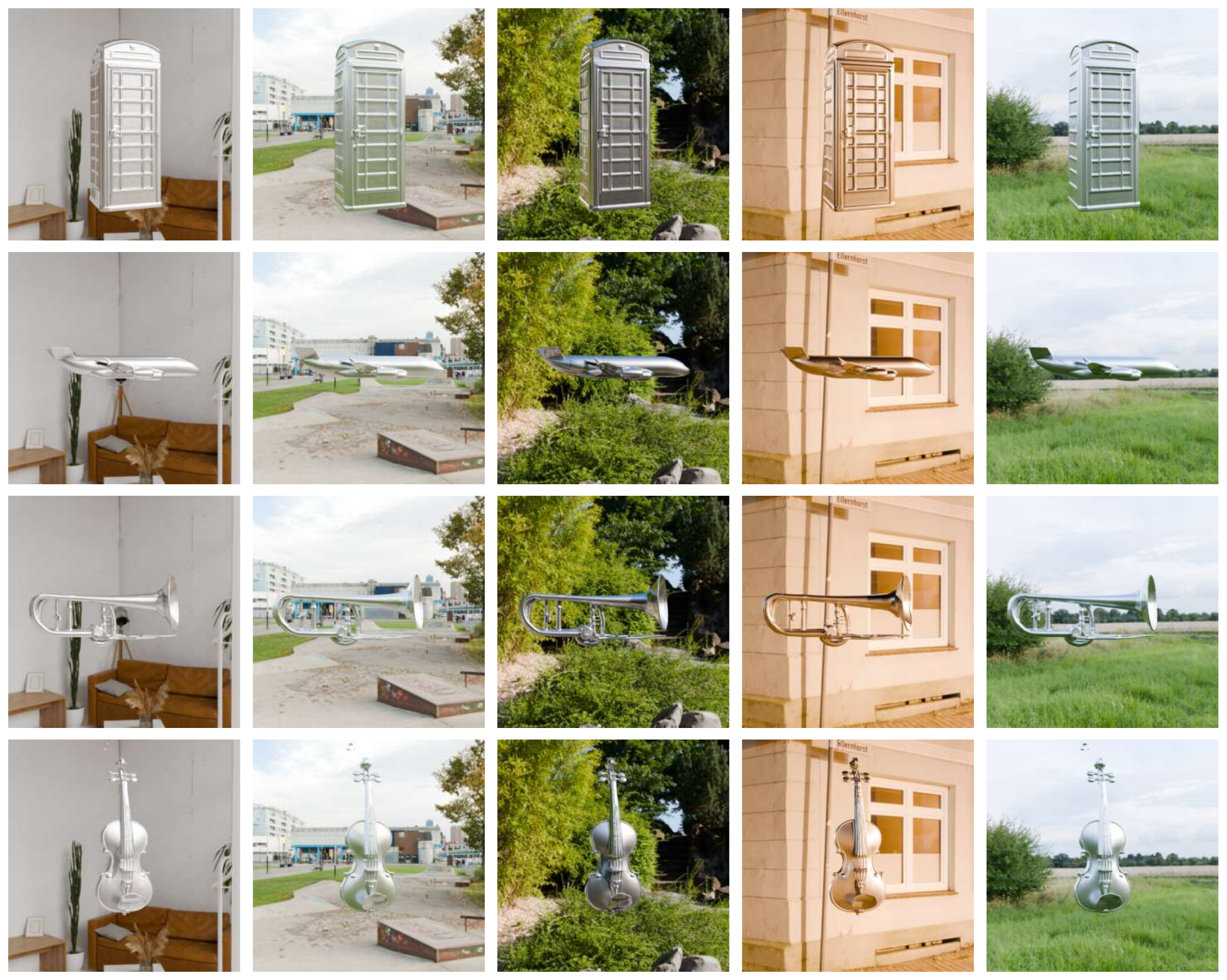}
    \caption{Generated 3D assets made of Steel material under various lighting condition}
    \label{fig:3dgen_lighting}
\end{figure*}

\fi

\end{document}